\documentclass[journal]{IEEEtran}

\usepackage{multirow}
\usepackage{graphicx}
\usepackage{textcomp}
\usepackage{amssymb}
\usepackage{subfigure}
\usepackage{amsmath}
  
\usepackage{color}
\usepackage[nocompress]{cite}
\newcommand{\djc}[1]{\textcolor{black}{#1}}

\ifCLASSINFOpdf
\else
\fi
\begin{document}
%
\title{Monocular Camera-based Complex Obstacle Avoidance via Efficient Deep Reinforcement Learning}
%
%
%

\author{Jianchuan~Ding\textsuperscript{$\dagger$},~Lingping~Gao\textsuperscript{$\dagger$},~Wenxi~Liu,~Haiyin~Piao,~Jia~Pan,~Zhenjun~Du,~Xin~Yang\textsuperscript{$*$},~Baocai~Yin
	\thanks{J. Ding is with the School of Computer Science, Dalian University of Technology, Dalian 116024, China, and also with Hebei University of Water Resources and Electric Engineering, Cangzhou, 061016, China. (e-mail:djc\_dlut@mail.dlut.edu.cn).}
	\thanks{L. Gao is with the School of Computer Science, Dalian University of Technology, Dalian 116024, China, and also with Alibaba Group, Hangzhou, 310000, China. (e-mail:gaolingping.glp@alibaba-inc).}
	\thanks{W. Liu is with the College of Mathematics and Computer Science, Fuzhou University, Fuzhou 350108, China (e-mail: wenxi.liu@hotmail.com).}
	\thanks{H. Piao is with Northwestern Polytechnical University, Xi'an 710072, China  (e-mail: haiyinpiao@mail.nwpu.edu.cn).}
	\thanks{J. Pan is with the Department of Computer Science, The University of Hong Kong, Hong Kong, China (e-mail: jpan@cs.hku.hk).}
	\thanks{Z. Du is with SIASUN Robot \& Automation Co., Ltd, Shenyang 110168, China (e-mail: duzhenjun@siasun.com).}
	\thanks{X. Yang and B. Yin are with Dalian University of Technology, Dalian 116024, China (e-mail: xinyang@dlut.edu.cn; ybc@dlut.edu.cn).}
	\thanks{\textsuperscript{$\dagger$} Equal contribution. \textsuperscript{$*$} Corresponding author.}
	\thanks{Copyright © 2022 IEEE. Personal use of this material is permitted. However, permission to use this material for any other purposes must be obtained from the IEEE by sending an email to pubs-permissions@ieee.org.}

}

\markboth{IEEE TRANSACTIONS ON CIRCUITS AND SYSTEMS FOR VIDEO TECHNOLOGY}%
{Shell \MakeLowercase{\textit{et al.}}: Bare Demo of IEEEtran.cls for IEEE Journals}

%



\maketitle

\begin{abstract}
	
Deep reinforcement learning has achieved great success in laser-based collision avoidance works because the laser can sense accurate depth information without too much redundant data, which can maintain the robustness of the algorithm when it is migrated from the simulation environment to the real world. However, high-cost laser devices are not only difficult to deploy for a large scale of robots but also demonstrate unsatisfactory robustness towards the complex obstacles, including irregular obstacles, \emph{e.g.}, tables, chairs, and shelves, as well as complex ground and special materials. In this paper, we propose a novel monocular camera-based complex obstacle avoidance framework. Particularly, we innovatively transform the captured RGB images to pseudo-laser measurements for efficient deep reinforcement learning. Compared to the traditional laser measurement captured at a certain height that only contains one-dimensional distance information away from the neighboring obstacles, our proposed pseudo-laser measurement fuses the depth and semantic information of the captured RGB image, which makes our method effective for complex obstacles. We also design a feature extraction guidance module to weight the input pseudo-laser measurement, and the agent has more reasonable attention for the current state, which is conducive to improving the accuracy and efficiency of the obstacle avoidance policy. Besides, we adaptively add the synthesized noise to the laser measurement during the training stage to decrease the sim-to-real gap and increase the robustness of our model in the real environment. Finally, the experimental results show that our framework achieves state-of-the-art performance in several virtual and real-world scenarios.
\end{abstract}
\vspace{-0.2cm}

%
\IEEEpeerreviewmaketitle

\vspace{-0.5cm}
\section{Introduction}
\vspace{-0.1cm}
%
%
%
%

\begin{figure}[t]
	\centering
	
	\includegraphics [height=125.04mm,width=89mm ]{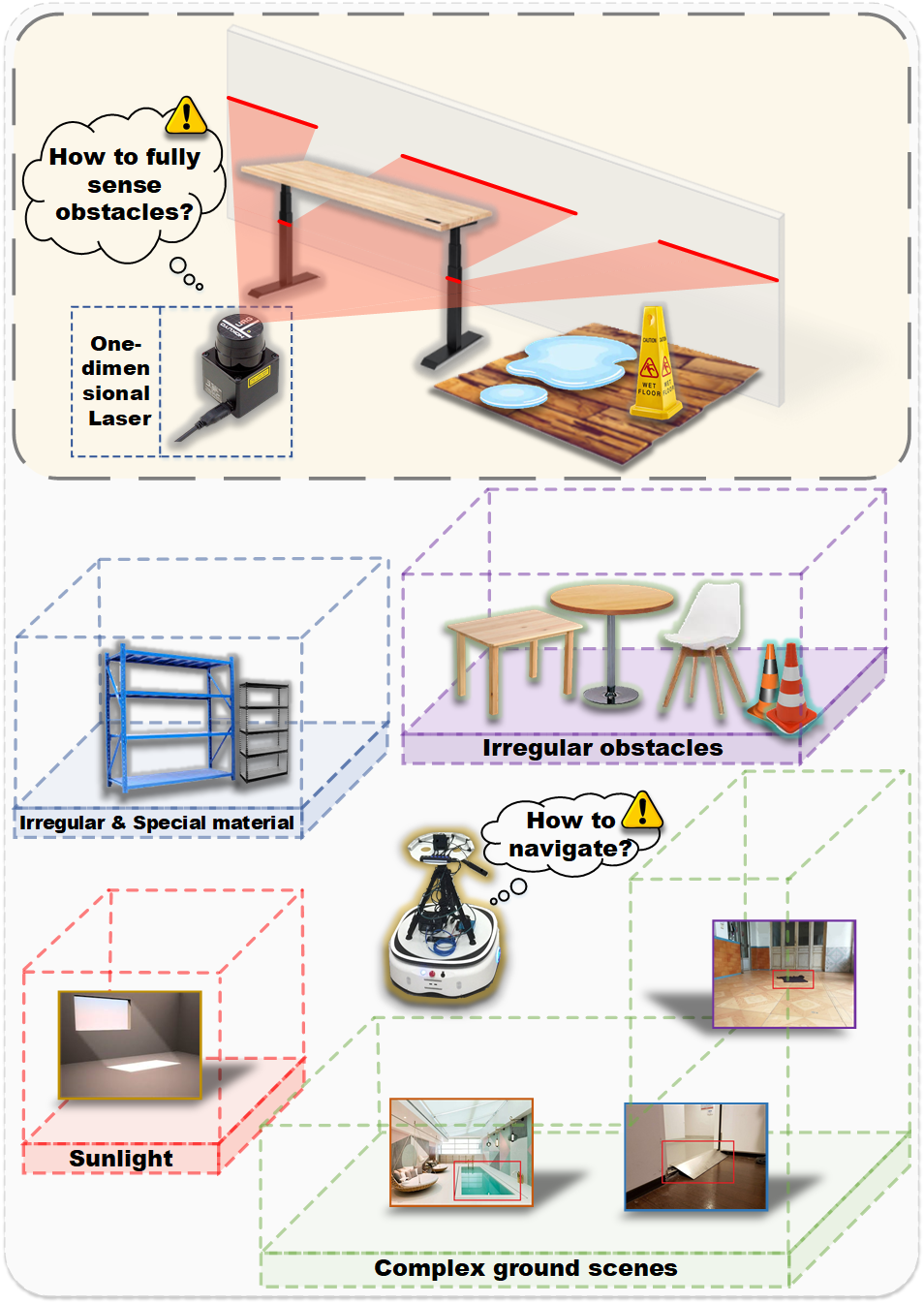}
	
	\caption{
		\djc{One-dimensional laser sensors have low robustness to the complex obstacles of certain types. As illustrated in the figure, they are (a) difficult to perceive the shape of irregular objects, (b) difficult to exploit the semantic information of complex ground scenes, and (c) also limited by some special materials. In addition, the influence of sunlight makes the depth of the RGB-D camera almost invalid.} 
	}
    
	\label{fig:teaser}
	\vspace{-0.4cm}
\end{figure}

\IEEEPARstart{C}{ollision} avoidance is a challenging research problem for robot navigation, which has been studied for decades and it can be applied in many important applications in real-world scenarios, \emph{e.g.}, unmanned driving~\cite{arokiasami2016interoperable} and freight logistics~\cite{schroeder2012towards}. In recent years, deep reinforcement learning (DRL) based methods~\cite{tai2016deep, long2018towards, choi2019deep, gao2021vision} have been studied to address the collision avoidance problem in the robot system. \djc{Compared with deep learning-based methods~\cite{pfeiffer2017perception, fan2018crowdmove, chen2017socially, zhang2020language} the DRL-based methods can collect a large amount of data based on simulation without relying on manually-labeled data.} Specifically, in virtual environments, the agent can be simulated to navigate, continuously interact with the environment in a `trial-error' manner, and obtain the environmental feedback for learning. Therefore, in practice, their collision avoidance policies can be trained in simulation and then migrated to real robots.

As the major limitation of learning policy using simulation data, the large gap between simulation and real-world makes it difficult to directly migrate the learned policy from virtual agents to real robots. In order to reduce this gap, one feasible solution is to apply lasers to sense the surrounding environment and formulate the observation as the obstacle distance information of discretized orientations in the form of one-dimensional vector~\cite{long2018towards}. Such laser-based observation possesses the advantage of accuracy and simplicity, while it not only ignores the discrepancy of texture and color between the simulation environment and the real environment, but also contains less redundant information than image data, which benefits the training of DRL frameworks. 

However, laser sensors notoriously have several shortcomings, \emph{i.e.}, one-dimensional laser observations cannot fully exploit the scene semantic information, and they are difficult to perceive the complex obstacles (as shown in Fig.~\ref{fig:teaser}, the irregular obstacles, the complex ground scenes, and special materials). Besides, its high price makes it unsuitable to deploy in large-scale industrial applications.

\djc{In this work, we aim to set up a collision avoidance framework for a real robot based on a monocular RGB camera using efficient DRL. Compared with laser sensors, RGB cameras are not only low-cost but also contain necessary semantic information for understanding the environment.}	
However, the image data captured by RGB cameras contains some redundant information~\cite{chen2019learning} (\emph{e.g.}, intensity, texture, \emph{etc}.), which causes a large sim-to-real gap that makes DRL difficult to converge and reduces the robustness of robots in unseen scenes.
\djc{Thus, to utilize the advantage of RGB cameras while reducing redundant information, we propose a novel pseudo-laser measurement generated from the single RGB image captured by a deployed monocular camera for the task of collision avoidance in the complex scene. Unlike~\cite{wang2019pseudo} directly mapping the depth map to Pseudo-LiDAR to improve the accuracy of 3D bounding box prediction, our pseudo-laser measurement fuses the distance and semantic information of the complex objects and the scene, which thus can more effectively handle the complex obstacles than the laser-based measurement. On the other hand, this pseudo-laser measurement is represented by one-dimensional data, which can provide more concise input for DRL to narrow the sim-to-real gap and promote the effective convergence of DRL and the generalization of the model.}
In particular, our approach for generating pseudo-laser measurement is composed of two stages: semantic masking and deep slicing.
We first estimate the scene depth map based on the RGB image captured by a monocular camera. 
It is reasonable to assume that the ground or floor depth that occupies a larger area in the depth map has no substantive significance for obstacle avoidance. Previous researchers~\cite{choi2019deep} usually used the method of projecting a depth map as a point cloud and then performing horizontal threshold-based cutting to remove the ground information. However, this method may fail in some complex ground scenes, \emph{e.g.}, water on the ground, clothes on the floor, and slopes (as shown in Fig.~\ref{fig:teaser}). Therefore, we exploit the semantic information of the scene to infer the traversable region, which locates these `special obstacles' (\emph{e.g.}, clothes, water, glass, \emph{etc}.) on the ground. We design the semantic mask to remove the distance information of the traversable region from the estimated depth map. In addition, we also extract the edge contours of obstacles and combine them with the later depth slicing module to improve the perception of complex obstacle shapes. In this way, the agent can fully exploit pixel-wise semantic information in the field of view (FOV), which is beneficial to make DRL efficient.

After culling out the traversable region from the depth map, we propose a depth slicing module to generate the pseudo-laser measurement. A naive way to accomplish pseudo-laser measurement is to slice the depth map along a horizontal axis. However, this approach ignores the shape of irregular obstacles. For instance, as shown in Fig.~\ref{fig:teaser}, if only slicing the feet of the table, the robot may collide with the upper part of it. Thus, we propose a dynamic local minimum pooling operation, which slices the depth information of each column from the semantic-depth map, performs minimum pooling for each column and composes them into a one-dimensional vector, \emph{i.e.} the pseudo-laser measurement, in which each element of the vector represents the distance away from the nearest obstacle along the vertical axis. 


With the pseudo-laser measurement, we adopt a DRL framework to sample the collision avoidance policy of the robot to efficiently avoid obstacles. In particular, to overcome the limitation of $90\,^{\circ}$-FOV and enhance the performance of the policy, we introduce the LSTM \cite{hochreiter1997long} submodule to leverage the historical state of the agent to model its temporal actions.
Then, we propose the Feature Extraction Guidance (FEG) module to improve the expressiveness of the pseudo-laser measurement. Generally, the state of the agent in DRL includes observation, goal, and velocity. There are two methods of state fusion as input to the DRL network: data-level fusion~\cite{choi2019deep} and decision-level fusion~\cite{long2018towards}. However, they are not conducive to process the input information from different modalities, so they result in unreliable DRL policies. Thus, we design the FEG module to fully exploit the interrelationship of the input information across the channel domain and the spatial domain. Specifically, FEG weights the pseudo-laser measurement according to the destination, distance, and velocity data, then the weighted result (feature mask) is producted with the observation data. It can pay more attention to the pseudo-laser measurement that guides the current obstacle avoidance task thus the subsequent DRL framework is more purposeful. Due to the preprocessing of the input, the initial DRL stage is not so blind, which accelerates the network convergence.

Finally, to optimize the training process of DRL, we design several simulation environments with different levels of difficulty. The algorithm starts training from simple scenes, then we migrate the trained model to more complex scenes. In order to transfer our model from the simulation environment to the real world, we augment the observations collected from the simulation scenes by adding specially-designed noises similar to the real world to the training data, thereby narrowing the sim-to-real gap.
We deploy 7 models in the Gazebo simulation environment, and the results show that our noise model has the highest success rate in complex environments. In addition, this model shows smooth motion and a high success rate in real scene experiments. Therefore, the data augmentation we adopted benefits the sim-to-real migration of our model.

The main contributions of our work are as follows:

\begin{itemize}
	\item We present a vision-based collision avoidance method based on deep reinforcement learning for robot navigation, which only relies on a single RGB monocular camera as the sensor. 
	
	
	\item \djc{To utilize the semantic information and share the advantage of laser sensors, we propose using RGB images to encode one-dimensional pseudo-laser measurement for collision avoidance, which greatly improves collision avoidance ability of robots in complex unknown scenes.}
	
	\item We design a feature extraction guidance module to improve the expressiveness of pseudo-laser measurement in the DRL process, thereby enhancing the ability of robots to understand and handle the obstacle avoidance task.
	
	\item To accommodate our vision-based policy, we introduce a new data augmentation to enhance the robustness of the model and reduce the sim-to-real gap. Experiments show that the robot moves smoothly through the adaptive data augmentation.
	
\end{itemize}

\section{Related Work}

	\subsection{Laser-based navigation}

Although the artificial potential fields \cite{cosio2004autonomous}, dynamic window approaches \cite{fox1997dynamic}, these traditional navigation methods can achieve safe obstacle avoidance in simple scenarios. However, the map information needs to be known, and it takes a lot of time or even fails due to the sensitivity to hyperparameters and local minima in some complex dynamic scenes. For example, the advanced traditional method ORCA \cite{van2011reciprocal} needs to know the location information of other agents and static map information. Its generalization is poor in complex scenes due to tedious parameter adjustment.

In recent years, with the development of deep learning, \cite{michels2005high, fan2018crowdmove, faust2018prm, chen2017socially} and \cite{fragkiadaki2015learning} have shown the effectiveness of learning-based navigation methods. Pfeiffer et al. \cite{pfeiffer2017perception} proposed a data-driven end-to-end motion planner. They used the data generated in the simulation environment through the ROS navigation package to train a model. It is able to learn the complex mapping from laser measurement and target positions to the required steering commands for the robot. This model can navigate the robot through a previously unseen environment and successfully react to sudden changes. Nonetheless, similar to the other supervised learning methods, the performance is seriously limited by the quality of the dataset. To solve the problem, Tai et al. \cite{tai2017virtual} proposed a mapless motion planner based on DRL to map 10-dimensional range findings to actions. The robot can safely navigate through a priori unknown environment without collision in this way. However, due to the sparse laser and the simple training scene, the model does not perform well on dynamic obstacles. \djc{Long et al. \cite{long2018towards} proposed a multi-scenario and multi-stage training framework based on the PPO \cite{schulman2017proximal} reinforcement learning method. By inputting three consecutive frames from a $180\,^{\circ}$ laser scanner, obstacle avoidance for dynamic objects is achieved, and it is extended to multi-robot applications. Fan et al.~\cite{fan2020distributed} integrated the learning algorithm of Long et al.~\cite{long2018towards} into the hybrid control framework to enable the agent to execute different decision models for different situations, further improving the robustness and effectiveness of the policy.} On this basis, Choi et al. \cite{choi2019deep} proposed an LSTM agent with Local-Map Critic (LSTM-LMC) to reduce the FOV to $90\,^{\circ}$ and achieve the performance of $180\,^{\circ}$. They used point cloud mapping and slicing to obtain laser data from the depth map. Due to the limitation of depth map slicing of point cloud mapping, the robot performs poorly on some complex ground (\emph{e.g.}, water surface, stairs, clothes, \emph{etc}.) and irregular obstacles. \djc{Tang et al.~\cite{tang2020reinforcement} proposed a spiking deep deterministic policy gradient (SDDPG) framework to train a LIF-based spiking actor network (SAN) for mapless navigation. This method combines spike neural network(SNN) with DRL to realize a low-power obstacle avoidance framework based on laser sensors. However, the training scenarios limit their ability to adapt to dynamic obstacles, and the one-dimensional laser sensor makes it difficult to deal with complex obstacles.} We continued the work of \cite{long2018towards,Tianyu2018} and reduced the FOV to $90\,^{\circ}$ in our way. And the robot has the ability to avoid obstacles to irregular objects through our pseudo-laser measurement.

\begin{figure*}[htbp]
	\centering
	
	\includegraphics [height=60mm,width=182mm ]{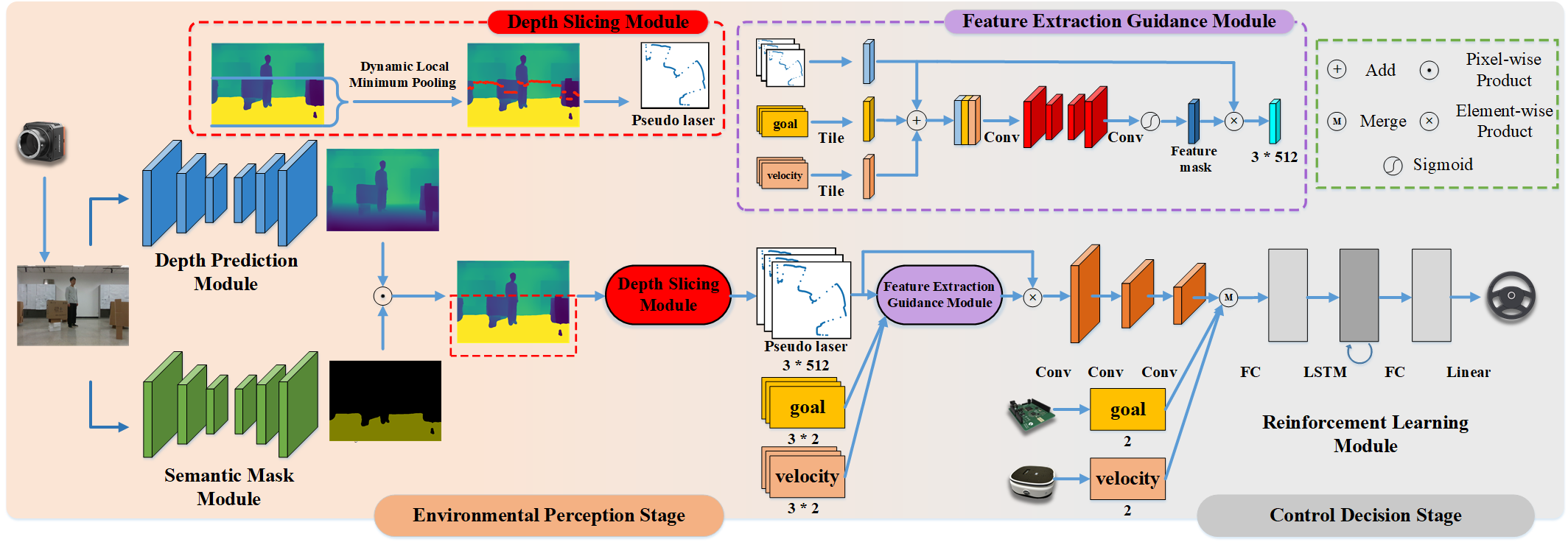}
	\caption{\djc{The pipeline of our entire framework. We fuse the distance and semantic information from the RGB image to generate pseudo-laser measurement and send them to the DRL network designed to obtain reasonable actions for navigating.}}
	\label{fig:pipeline}
	\vspace{-0.2cm}
\end{figure*}

\subsection{Vision-based navigation}

\djc{Due to laser scanners are expensive and can only capture limited information in the FOV. The camera can provide rich information about the operating environment of the robot and is low-cost, light-weight, and applicable for a wide range of platforms. There are a variety methods of vision-based navigation and obstacle avoidance, such as through optical flow \cite{souhila2007optical,mccarthy2004performance}, detection of vanishing points \cite{bills2011autonomous}, visual SLAM \cite{mur2015orb} and end-to-end mapping based on deep learning \cite{kim2015deep,giusti2015machine}. Nguyen et al.~\cite{nguyen2007real} used Sum of Absolute Differences (SAD)~\cite{huang2004global,vanne2006high,lengwehasarit2001probabilistic} correlation method to calculate the optimal disparity of stereoscopic cameras to develop an obstacle avoidance system for autonomous wheelchairs. However, traditional difference matching methods have difficulty in distinguishing complex obstacles on the ground.} Tai et al. \cite{tai2016deep} proposed an end-to-end model that can directly map depth maps to actions. However, operating a mobile robot to collect training data in the real world is inconvenient and time-consuming. \djc{\cite{ma2019learning,zhang2020language} and \cite{bharadhwaj2019data} trained policy for mapping from RGB image to action in the simulation environment. They perform well in training scenarios, but it is difficult to transfer to real-world environments. Wang et al. \cite{wang2021modular} achieved better results in robot navigation tasks by designing modular DRL methods from rewards and punishments. Although good results have been achieved in the simulator and simple real scenes, its state is expressed as a mixture of RGB images and laser data, which makes it difficult for its model to handle complex real scenes. (textures, colors, complex obstacles, \emph{etc.})} Gordon et al. \cite{gordon2019splitnet} used Habitat scene renderer \cite{savva2019habitat} to train visual navigation models, thereby reducing the gap between sim-to-real. \djc{On this basis, Lu et al.~\cite{lu2021mgrl} proposed a method combining graph neural network(GNN) and DRL to extract the structural relationship features of the scene, improve the learning efficiency of reinforcement learning, and further improve the accuracy of visual navigation. However, in unseen scenarios, these models need to be retrained. Similar works also include \cite{zhu2017target,wu2019exploring,yang2018visual,wu2020neonav} and \cite{wu2018building}.} Chen et al. \cite{chen2019learning} proposed a criterion for evaluating the gap between sim-to-real, and based on this proposed a visual obstacle avoidance method. However, due to the use of imitation learning and simple scenarios, the method is difficult to apply in challenging scenarios. Instead, we use a novel depth map slicing method that combines depth and semantic information to achieve mapping from a single RGB image to pseudo-laser measurement. The robot can safely avoid obstacles in more complex scenes in this way.

\subsection{Real-world random noise}

Owing to the over-idealization of sensors in the simulation environment, it is difficult to apply the policy trained in the virtual environment to real scenarios. To overcome this problem, \cite{tan2018sim,peng2018sim} and \cite{tobin2017domain} introduce random noise during the training process to simulate real sensor data. They showed that noise improves the robustness of the agents in challenging real-world. In this work, we also use data augmentation during training that simulates real scene depth slicing data to make the model better transfer to the real environment.


%

\section{Methodology}


In this paper, we propose the pseudo-laser measurement, which is generated by distance and semantic information from an RGB image, to improve the environmental-perception ability of robots for the efficient DRL policy. In addition, we design an FEG module and a data augmentation method to make the pseudo-laser measurement play a more reasonable role in the DRL module. Based on this, our method can effectively perceive irregular obstacles and complex ground information in the scene, and reasonably model the current observation, to perform efficient DRL and obstacle avoidance in complex scenes.

As shown in Fig.~\ref{fig:pipeline}, we obtain the corresponding depth map and semantic mask from the input RGB image. Then, we cull out the traversable region from the depth map through the semantic mask, since the depth of the traversable region under real conditions affects collision avoidance. After that, we propose a dynamic local minimum pooling operation that slices the semantic-depth map to generate the pseudo-laser measurement. Based on this, we input the pseudo-laser to the DRL network and sample the actions to guide the agent to avoid obstacles. In order to enable robots to capture the information that needs more reasonable attention in the input pseudo-laser, we design a FEG module. In the following, we first describe the method we proposed to generate pseudo-laser measurement, then introduce the DRL framework and submodules. Finally, we depict the data augmentation method of our model.


\djc{\subsection{Pseudo-laser measurement.}}
Our approach is inspired by the previous work~\cite{long2018towards} that uses the laser sensor to obtain the distance measurement away from the nearest obstacles and formulates them as a one-dimensional vector. The major advantage of this method is that they do not need to have the perfect sensing for the environment to perform collision avoidance, which significantly reduces the sim-to-real gap. 
\djc{The motivation of pseudo-laser measurement we proposed is to ignore the difference between simulation and reality. As simple and accurate one-dimensional data, pseudo-laser can also reduce the training difficulty of DRL while incorporating rich semantic information to improve robustness to complex obstacles.
Pseudo-laser measurement is fused through depth information and semantic information of the current state so that our method is robust in complex scenes. In particular, our approach for generating pseudo-laser measurement is composed of two stages: semantic masking and deep slicing.}

\textbf{\djc{Semantic-depth map \& Semantic mask module.}} We use the depth estimation model \cite{wofk2019fastdepth} and the image semantic mask model \cite{huang2019ccnet} as two parallel branches to obtain the depth map $M_D$ and semantic mask $M_G$ of the input RGB image $I$ ($I\in \mathbf{R}^{H\times W}$), respectively. 
\djc{The depth estimation branch employs pretrained MobileNet \cite{howard2017mobilenets} as the encoder and nearest-neighbor interpolation upsampling layers with the depthwise separable convolution in the decoder. For indoor and outdoor scenes, we train the depth estimation branch on the NYUv2 \cite{silberman2012indoor} and KITTI \cite{geiger2013vision} datasets, respectively. Intuitively, since the information of the traversable area (\emph{e.g.}, ground) cannot help the robot avoid obstacles, we need to delete the depth of the traversable area in the estimated depth map to avoid its impact on the perception of the environment.} 

\begin{itemize}
	\item \textbf{Traversable region} - Areas outside of any geometric and non-geometric substances that will affect robot navigation and obstacle avoidance.
\end{itemize}

To achieve this purpose, we leverage the second branch to produce a semantic mask providing pixel-level traversable region labels. \djc{Since the only concern at this stage is `which pixel in the image belongs to the traversable region', our second branch can provide a binary mask $M_G$, indicating which pixel of the image belongs to the traversable region, \emph{i.e.}, $M_G(i,j)=0$ if the pixel at $(i,j)$ belongs to the traversable region, otherwise $M_G(i,j)=1$.}
Due to the useless depth of traversable region for collision avoidance task, we generate a depth map without traversable region depth via $M = M_D \odot M_G$, where $\odot$ refers to the pixel-wise multiplication. \djc{Therefore, the semantic mask $M_G$ obtained by the Semantic Mask Module can effectively handle complex ground scenes. (Fig.~\ref{fig:pipeline} shows more details)}

\djc{\textbf{Depth slicing module.}} In order to obtain the pseudo-laser measurement $L$, the naive solution is to slice the depth map along a specific row, \emph{i.e.}, $L = M(i,:)$, where $i$ indicates the index of the row and thus $L\in \mathbf{R}^{1\times W}$. However, the naive solution may not be robust for irregular obstacles. \djc{For instance, if the height of slicing is on the feet of the table, the robot is likely to collide with the upper part of the table. When there exists a small obstacle, such a naive slicing method may cause a similar problem. To address this problem, we propose an efficient solution, \emph{i.e.}, dynamic local minimum pooling operation. Specifically, we only keep the lower half of the depth map for computation, \emph{i.e.}, $\hat{M} = M(\frac{H}{2}:H, :)$, as shown in Fig.~\ref{fig:pipeline} Depth Slicing Module. This is because the upper half of the image often provides little context information for collision avoidance.} Then, we perform column-wise minimum pooling to select the minimal value along each column, \emph{i.e.}, the pseudo-laser measurement $L$ can be computed as $L(j)=\min \hat{M}(:,j), \forall j=\{1,\cdots, W\}$), where $\hat{M}(i,j) \ne 0$. \djc{Intuitively, the minimum pooling operation approximately localizes the nearest obstacles of any height in the environment. Hence, the extracted pseudo-laser measurement with spatial semantic information about the scene are forwarded to the DRL framework.}


\subsection{Deep reinforcement learning module.}
In this part, we hope that agents can effectively avoid obstacles in complex unknown environments and rely on vision only to safely reach the location of a man-made designated target.

\textbf{DRL problem setting.} Affected by the FOV of the sensor, the agent cannot fully perceive the surrounding environment during the training process, thus we define the training process as a partially observable Markov decision process (POMDP).
In simple terms, a POMDP is a cyclical process in which an agent takes action to change its state to obtain rewards and interact with the environment. 



Because the action space of robots is continuous, we adopt the Actor-Critic (AC) framework based on policy gradient to implement our DRL obstacle avoidance policy.

\begin{itemize}
	\item \textbf{Observation space.}
	For the observation $o$ of the agent, we use the pseudo-laser measurement with $90\,^{\circ}$ horizontal FOV. 


\item \textbf{Action space.}
\djc{For the action $a$ of the agent, we use R$^{2}$ vectors for linear($v$) and angular($w$) velocities. Linear velocity of the agent is in range $v \in (0.0,1.0)$m/s and angular velocity is in range $w \in (-90.0,90.0)\,^{\circ}$/s. We use normalized angular velocity which is in the range $w \in (-1.0,1.0)$ as outputs of the neural networks. Note that the robot cannot move backward (\emph{i.e}. $v < 0.0$m/s) because the laser sensor cannot cover the back area of the robot.}

\item \textbf{Reward function.}
Our objective is to avoid collisions during navigation and minimize the mean arrival time of robots. The reward function follows \cite{long2018towards}. 
\end{itemize}

\djc{
\begin{equation} 
r_{i}^{t} = (r_{goal})_{i}^{t}+(r_{collision})_{i}^{t}+(r_{rotational})_{i}^{t} 
\vspace{-0.1cm}
\end{equation}  
}
\djc{
The total reward $r$ received by robot $i$ at timestep $t$ is a sum of three terms, $r_{goal}$, $r_{collision}$, $r_{rotational}$.
$r_{goal}$ represents the reward of whether the agent has reached its goal. If the agent collides, $r_{collision}$ is a penalty. And $r_{rotational}$ encourages the agent to move smoothly.
}
\djc{
\begin{small}
	\begin{equation}
	(r_{goal})_{i}^{t} = \left\{ \begin{array} { l l } { r _ { arrival } } & { \text { if } \| \textbf{p} _ { i } ^ { t } - \textbf{g} _ { i } \| < 0.1 } \\ { \omega _ { g } ( \| \textbf{p} _ { i } ^ { t-1 } - \textbf{g} _ { i } \|-\| \textbf{p} _ { i } ^ { t } - \textbf{g} _ { i } \| ) } & { \text { otherwise } } \end{array} \right.
	\end{equation}
\end{small}
}
\djc{
${\textbf{p} _ { i } ^ { t }}$ is the position of robot $i$ at time $t$ and $\textbf{g} _ { i }$ is the goal position of robot$i$.
In order to avoid collisions between agent and obstacles, we set $r_{collision}$ as a penalty.
}
\djc{
\begin{equation}
(r_{collision})_{i}^{t} = \left\{ \begin{array} { l l } { r _ {  collision} } & { \text { if } \| \textbf{p} _ { i } ^ { t } - \textbf{p} _ { j } ^ { t } \| < 2 R } \\ { } & { \text { or } \| \textbf{p} _ { i } ^ { t } - \textbf{B} _ { k } \| < R } \\ { 0 } & { \text { otherwise } } \end{array} \right.
\end{equation}
}
\djc{
$\textbf{B} _ { k }$ is the position of obstacle $k$ and $R$ is the radius of the robot. 
To encourage the robot to move smoothly, a small penalty $(r_{rotational})_{i}^{t}$ is introduced to punish the large rotational velocities.
}

\djc{
\begin{equation}
(r_{rotational})_{i}^{t} = \omega _ { w } | w _ { i } ^ { t } | \quad \text { if } | w _ { i } ^ { t } | > 0.7
\end{equation}
}
\djc{
We set $r _ { arrival }$ = 15, $\omega _ { g }$ = 2.5, $r _ {  collision}$ = -15 and $\omega _ { w }$ = -0.1 in the training procedure.
}

\djc{
Our DRL framework is based on the module of \cite{long2018towards}. Compared with their method, the FOV of our framework is much smaller (\emph{i.e.}, $90\,^{\circ}$) and the observation may be noisier. In order to make up for the limitation of the sensor FOV, we introduce LSTM into the decision policy network. At the same time, we design a FEG module to weight the input pseudo-laser measurement to further improve the ability of robots to understand and capture the surrounding environment.
}

\textbf{LSTM module.} As shown in Fig.~\ref{fig:pipeline}, we introduce the LSTM into our network to solve the limited FOV problem. LSTM has been applied as a temporal model to address the long-term dependency issue \cite{hochreiter1997long}. Here, LSTM allows the agent to make more reasonable actions based on the current state and the historic states. In specific, we add LSTM behind the fully connected layer where the behavioral state (\emph{i.e.}, the goal $\textbf{g}$ and velocity $\textbf{v}$ of the robot) and the observation \emph{O} are merged. The input of LSTM is a 256-dimensional feature and the output is also a 256-dimensional feature. Finally, the feature will be passed through another fully-connected layer and to map an action.

\textbf{\djc{Feature extraction guidance module.}}
In the DRL network, we take three consecutive pseudo-laser measurement ($L_{t-2}$, $L_{t-1}$, $L_{t}$), goal ($G_{t-2}$, $G_{t-1}$, $G_{t}$) and velocity ($V_{t-2}$, $V_{t-1}$, $V_{t}$) as input. Such input can enable our agents to infer the motion of surrounding objects, and thus make more effective collision avoidance behaviors. However, the different times and positions of the goal and velocity data in the pseudo-laser measurement have different effects on the current obstacle avoidance of the agent. Data representation of different modality probably has hierarchical attention in the pseudo-laser measurement. In order to make the agent better capture the reasonable information in the pseudo-laser measurement to perform efficient DRL, we introduce a FEG module to predict a mask to weight the pseudo-laser measurement. Specifically, as shown in the feature extraction guidance module of Fig.~\ref{fig:pipeline}, velocity vectors ($\mathbf{R}^{2}$) and goal vectors ($\mathbf{R}^{2}$) are tiled to match the shape of the pseudo-laser vector ($\mathbf{R}^{d_{laser}\ast2}$) where d$_{laser}$ is the size of the pseudo-laser vector. Then, these tiled vector is concatenated to the pseudo-laser vector ($\mathbf{R}^{d_{laser}\ast1}$), resulting in a matrix that has a size of $\mathbf{R}^{d_{laser}\ast5}$. We stacked these matrices from 3 recent timesteps to obtain a network input tensor ($\mathbf{R}^{d_{laser}\ast5\ast3}$). \djc{This tensor first passes through several convolutional layers and deconvolution layers whose activation function is RELU, and finally gets the final mask M$_{G}$ ($\mathbf{R}^{d_{laser}\ast3}$) through the Sigmod activation function. After obtaining the mask, we multiply the mask with the pseudo-laser measurement to obtain the final weighted pseudo-laser measurement. Finally, the weighted pseudo-laser measurement can be used as an input to the DRL network, and the obstacle avoidance action of the agent is returned. The DRL model can pay attention to laser data in the goal direction, in the velocity direction, and in close proximity, by weighting the pseudo-laser measurement. (See Sec.~\ref{sec:exp} for details.)}

\textbf{Training.}
Similar to the previous work~\cite{long2018towards}, our training is conducted on a virtual environment, in which agents interact with the environment and receive rewards from the environment to guide their learning. Since the DRL algorithm performs backpropagation based on the reward given by the interactive environment, if the environment is too complex, or the task is too difficult, the environment will feedback a sparse reward to the agent, which will cause the DRL framework to converge too slowly. In order to improve the efficiency of algorithm training, following the multi-stage multi-scenario training strategy~\cite{long2018towards}, we first train our DRL framework in simple scenarios for simple tasks and then transfer the trained model to a more complex environment for further training.

\begin{figure}[t]
	\centering
	\vspace{-0.2cm}
	\includegraphics [height=40mm,width=86mm ]{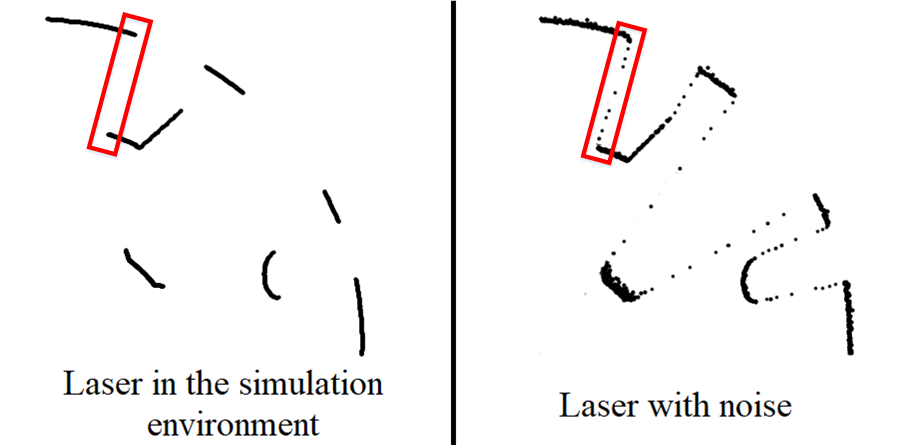}
	\caption{\djc{The raw laser measurement captured in the virtual scene and the noisy laser measurement after our data augmentation.}}
	\label{fig:noise}
	\vspace{-0.2cm}
\end{figure}

\subsection{Data augmentation.}
To enhance the robustness of the DRL based model in real-world scenarios, we add specific noise to the training laser measurement collected in the simulation environment to reduce the sim-to-real gap. 
In real-world scenes, the measurement errors of observations often occur around the boundaries of objects if some parts of one object block the other, \emph{i.e.}, the depth values of the boundary of the blocking part may be uncertain. Therefore, we intend to add noise to emulate such measurement errors in order to produce adversarial data during training.

To identify the junction boundary from the training laser measurement, \emph{i.e.}, a one-dimensional vector with precise distance information, we assume that, if the difference of two adjacent values in the vector is larger than the threshold $\alpha$, there may exist a junction boundary. To add noise, we locate the neighborhood of the junction boundary in the vector and replace all the values within the neighborhood by linearly interpolating the endpoints of the neighborhood (see Fig.~\ref{fig:noise}). 
\djc{On the other hand, for each vector element outside the neighborhood of the junction boundary, we adaptively add Gaussian white noise whose variance is proportional to the vector value. The policy can accommodate the sim-to-real gap of pseudo-laser measurement by introducing this data augmentation.}

\begin{figure}[b]
	\centering
	\vspace{-0.2cm}
	\includegraphics [height=54.22mm,width=70mm ]{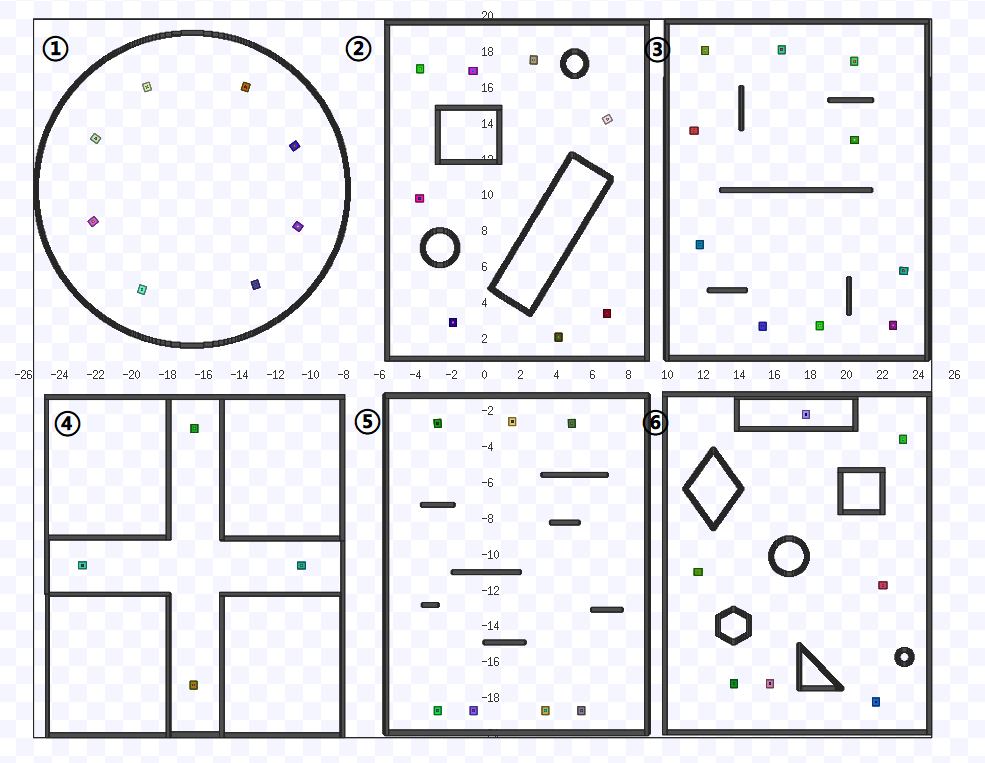}
	\caption{The training and testing scenarios built for the DRL framework in the Stage simulation environment. (1)-(3) are used for training. (4)-(6) are used for testing.}
	\label{fig:stage}
	\vspace{-0.2cm}
\end{figure}

\section{Experimental Results}
\label{sec:exp}

	In this section, we first test the DRL framework and evaluate its performance. We also show the visualization results of the FEG module. Then, we conduct experiments on the entire monocular camera-based collision avoidance method. Finally, we apply our model to a robot in real-world scenarios.

\begin{figure}[t]
	\centering
	
	\includegraphics [height=35.197mm,width=89mm ]{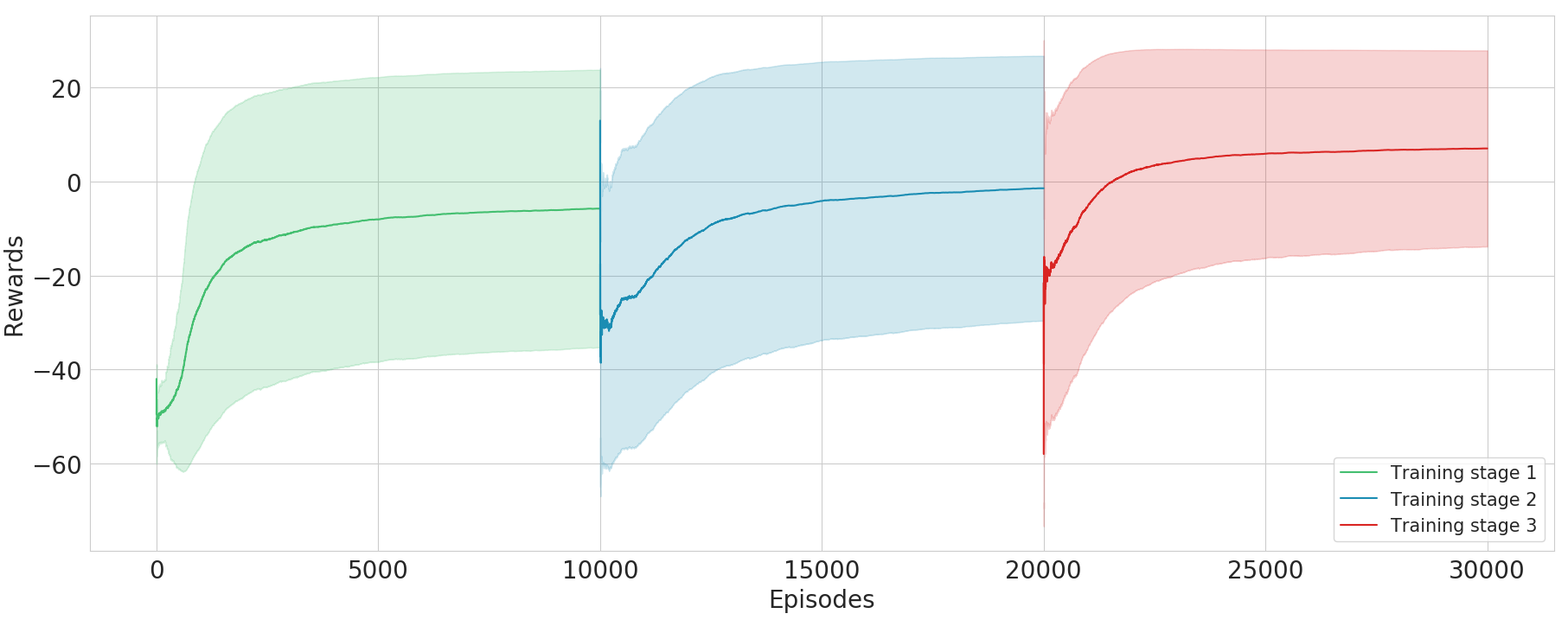}
	\caption{The average (dark curve) and standard deviation (light area) of rewards during the three stages of training episodes.}
	\label{fig:convergence}
	\vspace{-0.0cm}
\end{figure}

\textbf{Implementation details.}
Our algorithm is implemented in Pytorch. We train the DRL policy on a computer equipped with an i7-7700 CPU and NVIDIA GTX 1080Ti GPU for approximately 40 hours. We build training and testing scenarios based on the Stage mobile robot simulator (as shown in Fig.~\ref{fig:stage}) and train 4 models with hyperparameters (see Table~\ref{para}) while adopting a multi-scene multi-stage alternating training method. During training, the convergence of our reward function is shown in Fig.~\ref{fig:convergence}. Specifically, we train our model in three training scenes (see Fig.~\ref{fig:stage}) one after another. Note that we use a multi-agent collision avoidance task where each agent is assigned a shared collision avoidance policy, which can increase the robustness and training speed of our model.
	
\newcommand{\tabincell}[2]{\begin{tabular}{@{}#1@{}}#2\end{tabular}}

\begin{table}[t]	
	\setlength{\abovecaptionskip}{-0.2cm}
	\setlength{\belowcaptionskip}{-1.0cm}
	\caption{Hyper parameter settings in our experiments.}
	\label{para}
	\begin{center}
		\begin{tabular}{|c|c|}
			
			\hline
			\textbf{Hyper parameter} & \textbf{Value} \\
			\hline
			Batch size & 1024 \\
			\hline
			Maximum time steps & 150 \\
			\hline
			Training episode & 1,500,000 \\
			\hline
			Discount factor $\gamma$ & 0.99 \\
			\hline
			Learning rate & 5e-5 \\ 
			\hline
			LSTM unroll & 20 \\
			\hline
			Target network update ratio & 0.01 \\
			\hline
			KL penalty coefficient & 15e-4 \\
			\hline
			the noise threshold $\alpha$ & 0.5 \\
			\hline
		\end{tabular}
	\end{center}
\end{table}

We use the following metrics to evaluate the performance of different navigation algorithms:

\begin{itemize}
	
	\item \textbf{Success Rate.} The number of times a robot reaches the target waypoint without collision and overtime.
	
	\item \textbf{Average Time.} The average time it takes for a robot to reach the goal safely. 
	
\end{itemize}

\begin{figure}[hb]
	\centering
	\vspace{-0.2cm}
	\includegraphics [height=50.85mm,width=70mm ]{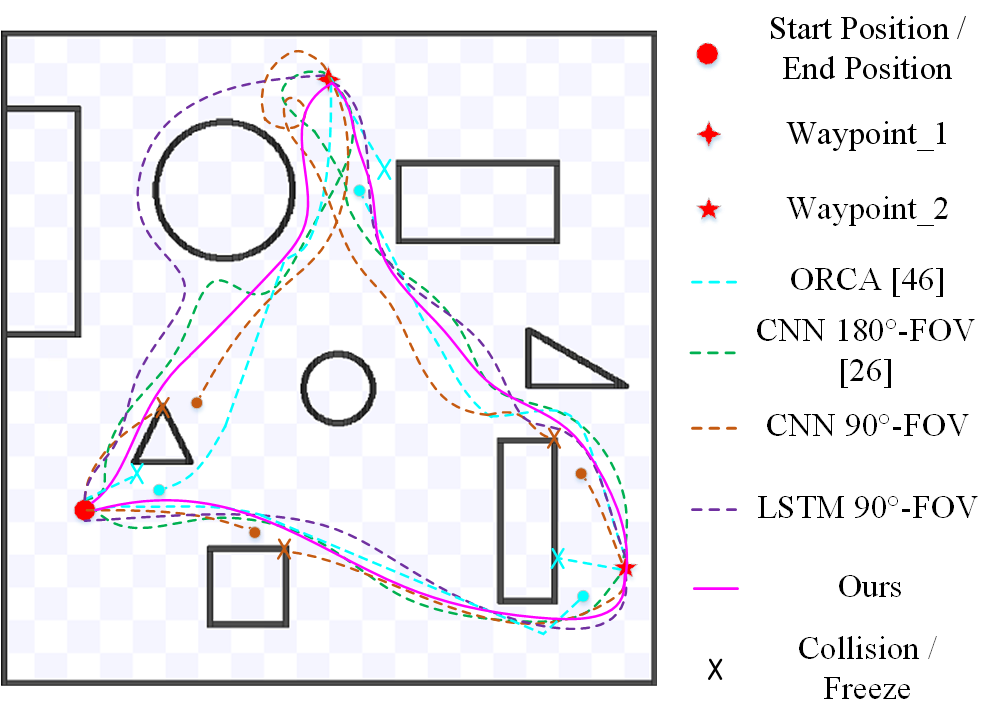}
	\caption{The trajectories generated by different models during navigating an agent in a virtual scene with two waypoints.}
	\label{fig:traj}
	\vspace{-0.0cm}
\end{figure}

\begin{figure*}[t]
	\centering
	\subfigure[Current state of the robot]{
		\begin{minipage}[t]{0.24\linewidth}
			\centering
			\includegraphics[height=46mm, width=35mm]{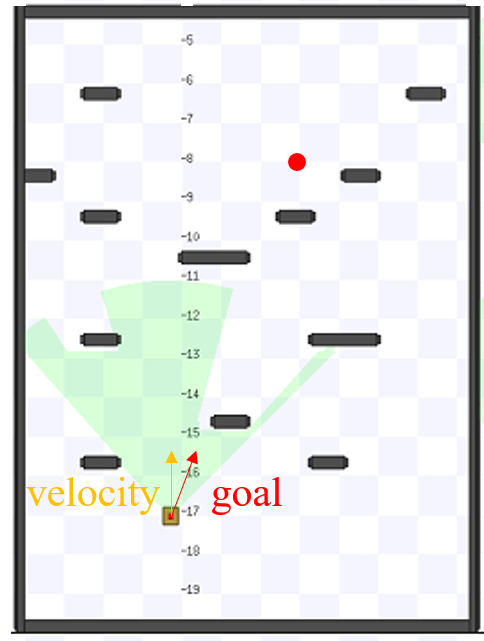}
			\label{fig:current}
		\end{minipage}%
	}%
	\subfigure[Laser input]{
		\begin{minipage}[t]{0.24\linewidth}
			\centering
			\includegraphics[height=43mm, width=42mm]{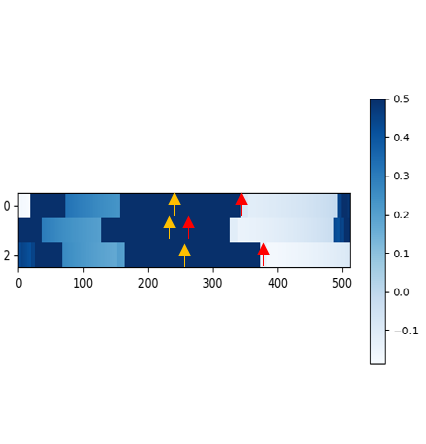}
			\label{fig:input}
		\end{minipage}%
	}%
	\subfigure[Feature mask]{
		\begin{minipage}[t]{0.24\linewidth}
			\centering
			\includegraphics[height=43mm, width=42mm]{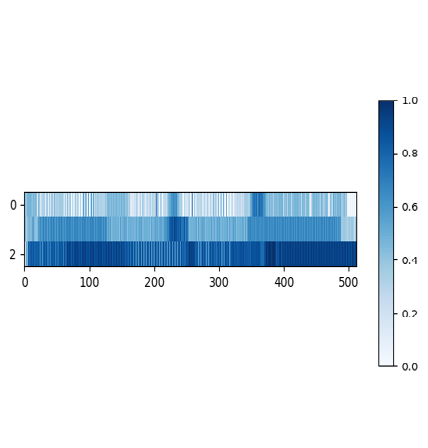}
			\label{fig:mask}
		\end{minipage}
	}%
	\subfigure[Fusion result]{
		\begin{minipage}[t]{0.24\linewidth}
			\centering
			\includegraphics[height=43mm, width=42mm]{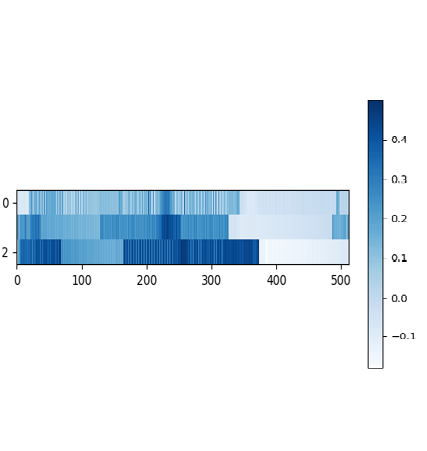}
			\label{fig:output}
		\end{minipage}
	}%
	\centering
	\caption{\djc{Visualization results of the FEG module mask in (a) state. The red arrow is the current relative direction to the goal, the yellow arrow is the current velocity direction, and the red dot is the goal position.}}

\end{figure*}

\textbf{\djc{Network structure experiment.}} \djc{To demonstrate that the Control Decision Stage of the $90\,^{\circ}$-FOV is reliable, we evaluate 4 different network structures (CNN models with $180\,^{\circ}$ and $90\,^{\circ}$ FOV, LSTM based model and LSTM+Attention model with $90\,^{\circ}$-FOV) in DRL models relying on laser sensors and ORCA \cite{van2011reciprocal} as baseline in the test scenarios shown in Fig.~\ref{fig:stage}. Note that the CNN model with $180\,^{\circ}$-FOV is \cite{long2018towards}, while the CNN model with $90\,^{\circ}$-FOV is the same model yet trained with sensor observation of $90\,^{\circ}$-FOV.}
The test scenario includes the collision avoidance tasks against static/dynamic obstacles: crossing, through walls, through fixed obstacles, and multi-robot obstacle avoidance, which are challenging scenes for evaluating the collision avoidance ability of our model against different types of obstacles. We randomly assign an initial position and a goal position to each robot and test it 100 times in different scenarios. The results are shown in Table~\ref{example_stage}. We also visualize the example trajectories of each model, as shown in Fig.~\ref{fig:traj}. Compared with peer models, our model can generate a smooth trajectory that successfully passes through two waypoints and reaches the goal without getting stuck or collisions.

\begin{table}[ht]
	\scriptsize	
	\setlength{\abovecaptionskip}{-0.2cm}
	\setlength{\belowcaptionskip}{-1.0cm}
	\caption{\scriptsize PERFORMANCE OF AGENTS WITH VARIOUS FOV AND ARCHITECTURES}
	\label{example_stage}
	\begin{center}
		\begin{tabular}{|c|c||c|c|c|c|c|}
			
			\hline
			\multicolumn{2}{|c||}{Model} & \tabincell{c}{ORCA \\ \cite{van2011reciprocal}} & \tabincell{c}{CNN \\ \cite{long2018towards}} & CNN & LSTM & \tabincell{c}{LSTM \\ + FEG \\ (Ours)}\\
			\hline
			\multicolumn{2}{|c||}{Field of View (FOV)} & - & $180\,^{\circ}$ & $90\,^{\circ}$ & $90\,^{\circ}$ & $90\,^{\circ}$\\
			\hline
			\multirow{2}{*}{Scene 4} 
			& \tabincell{c}{Success \\ rate} & 92\% & \textbf{98\%} & 79\% & 87\% & 97\% \\
			\cline{2-7}
			& \tabincell{c}{Average \\ time} & 7.516 & \textbf{6.331} & 8.035 & 6.589 & 6.989 \\
			\hline
			\multirow{2}{*}{Scene 5} 
			& \tabincell{c}{Success \\ rate} & 72\% & 86\% & 16\% & 78\% & \textbf{87\%} \\
			\cline{2-7}
			& \tabincell{c}{Average \\ time} & 16.844 & \textbf{10.882} & 23.549 & 11.752 & 12.954 \\
			\hline
			\multirow{2}{*}{Scene 6} 
			& \tabincell{c}{Success \\ rate} & 78\% & \textbf{95\%} & 63\% & 80\% & \textbf{95\%} \\
			\cline{2-7}
			& \tabincell{c}{Average \\ time} & 18.551 & \textbf{13.563} & 25.662 & 14.216 & 15.013 \\
			\hline
		\end{tabular}
	\end{center}
\end{table}

We observe that our method is superior to the collision avoidance method ORCA \cite{van2011reciprocal} and comparable to the model with $180\,^{\circ}$-FOV, while our model has a better success rate. For the $90\,^{\circ}$-FOV model, the success rate drops significantly. With the LSTM module introduced into the model, the performance of the model in a limited FOV is improved. Based on this, we propose the FEG module that CNN has reasonable attention to the current observation. Intuitively, it can make the policy reach or even exceed the model with $180\,^{\circ}$-FOV. However, due to the complexity of the network structure, the robot may occasionally decelerate significantly in dangerous situations, resulting in an increase of average time.

\begin{figure}[t]
	\centering
	
	\includegraphics [height=55mm,width=81mm ]{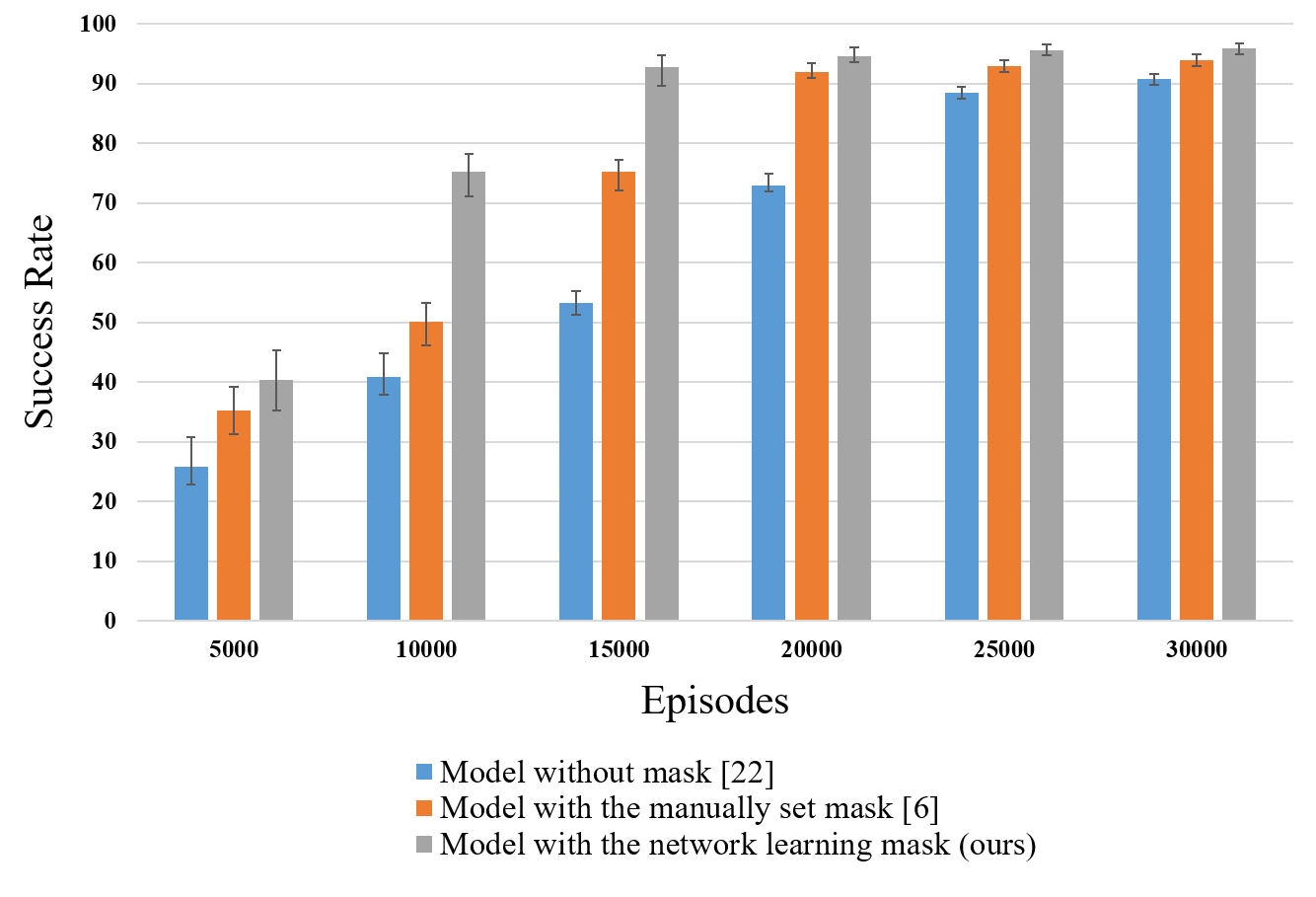}
	\caption{The success rate of the three models at different times illustrates the convergence rate of the different models.}
	\label{fig:convergence speed}
	\vspace{-0.0cm}
\end{figure}

\textbf{Convergence speed experiment.}
We find that after introducing the FEG module, the convergence speed of the algorithm is greatly accelerated, and the policy can achieve a better effect in a shorter time. Therefore, we conduct a comparative experiment of convergence speed. Specifically, we compare the three models (model without mask \cite{long2018towards}, model with the manually set mask \cite{choi2019deep}, model with the network learning mask), and train a fixed number of episodes for each model. The value of episodes represents the training time. We conduct a test navigation success rate for different training time and different model policies. As shown in Fig.~\ref{fig:convergence speed}, the results show that our model has the fastest convergence rate, with a success rate of 75\% at 10000 episodes, and a success rate of more than 90\% at 15000 episodes. The model without mask only reaches a 90\% success rate at 25000 episodes, showing a steady growth trend. \djc{Although the manually set mask has a faster convergence rate than the model without the mask, the effect is not ideal compared with the model where we obtain the mask through learning. In the same training times, our FEG module always maintains the highest success rate. Intrinsically, by obtaining masks that have hierarchical attention to the current observation, the network can more efficiently extract reasonable information for obstacle avoidance tasks and accelerate policy convergence.}

\textbf{\djc{Feature extraction guidance module analysis.}}
To study the internal effect of the FEG module, we show the visualization results of the learned mask. Fig.~\ref{fig:current} shows the current state of the robot, green represents the laser range, the red arrow is the current relative direction to the goal, the yellow arrow is the current velocity direction, and the red dot is the goal position. \djc{Fig.~\ref{fig:input} to~\ref{fig:output} are the visualization of laser input, feature mask and fusion result. Note that the visualization result is divided into three layers from top to bottom, corresponding to three consecutive times of laser measurement. We observe that the weight of the FEG module mask gradually increases from top to bottom. The network has different attention to different moments and pays more attention to the data at the current moment. Mask is obtained by fusion of laser input, goal, and velocity. Therefore the laser measurement in the two directions of the current position to goal direction and speed direction are given higher weight. The position of the obstacle (small value) in the laser input has a higher weight, thus the network pays more attention to the obstacles at closer distances, while the safety distance pays less attention. }

\begin{table}[t]
	\scriptsize	
	\setlength{\abovecaptionskip}{-0.2cm}
	\setlength{\belowcaptionskip}{-1.0cm}
	\caption{\scriptsize PERFORMANCE OF AGENTS WITH VARIOUS FOV AND ARCHITECTURES IN DENSE CROWD ENVIRONMENT}
	\label{dense crowd}
	\begin{center}
		\begin{tabular}{|c|c||c|c|c|c|c|}
			
			\hline
			\multicolumn{2}{|c||}{Model} & \tabincell{c}{ORCA \\ \cite{van2011reciprocal}} & \tabincell{c}{CNN \\ \cite{long2018towards}} & CNN & LSTM & \tabincell{c}{LSTM \\ + FEG \\ (Ours)}\\
			\hline
			\multicolumn{2}{|c||}{Field of View (FOV)} & - & $180\,^{\circ}$ & $90\,^{\circ}$ & $90\,^{\circ}$ & $90\,^{\circ}$\\
			\hline
			\multirow{2}{*}{Scene 11} 
			& \tabincell{c}{Success \\ rate} & 66\% & \textbf{72\%} & 24\% & 63\% & 69\% \\
			\cline{2-7}
			& \tabincell{c}{Average \\ time} & 105.764 & 95.056 & 121.655 & 98.432 & \textbf{89.535} \\
			\hline
			\multirow{2}{*}{Scene 12} 
			& \tabincell{c}{Success \\ rate} & 61\% & \textbf{70}\% & 39\% & 64\% & 68\% \\
			\cline{2-7}
			& \tabincell{c}{Average \\ time} & 202.766 & \textbf{176.631} & 215.906 & 194.024 & 180.787 \\
			\hline
			\multirow{2}{*}{Scene 13} 
			& \tabincell{c}{Success \\ rate} & 71\% & 75\% & 52\% & 68\% & \textbf{77\%} \\
			\cline{2-7}
			& \tabincell{c}{Average \\ time} & 357.803 & \textbf{296.438} & 349.970 & 321.009 & 306.556 \\
			\hline
		\end{tabular}
	\end{center}
	\vspace{-0.2cm}
\end{table}

\begin{figure}[t]
	\centering
	\vspace{-0.2cm}
	\includegraphics [height=71.64mm,width=75mm ]{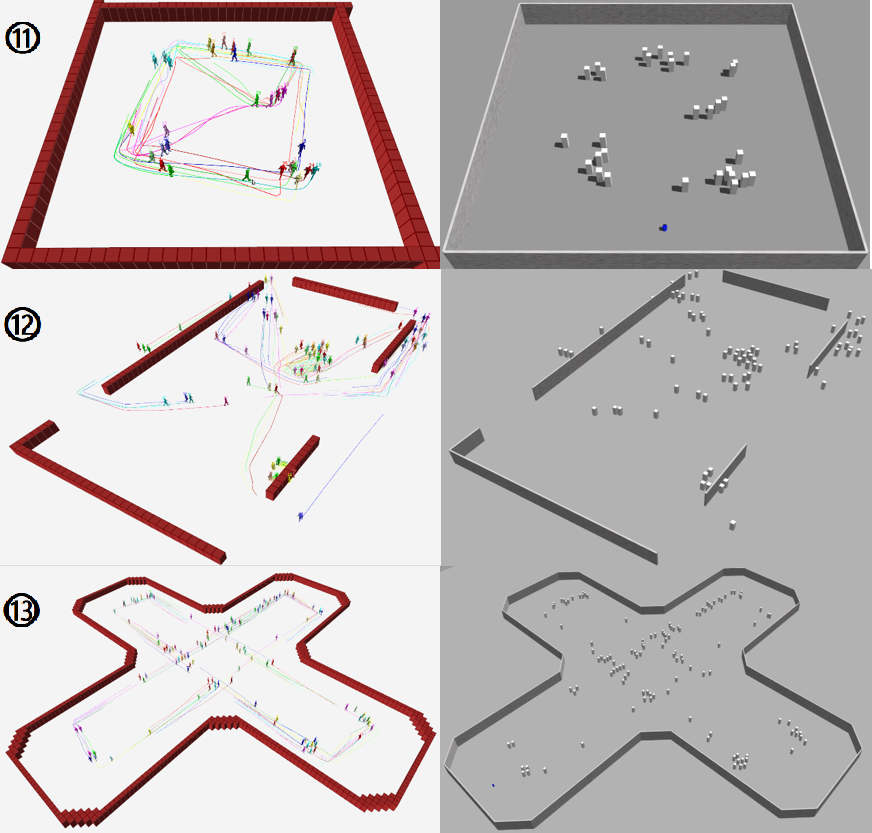}
	\caption{The virtual scene of the dense crowd experiment, the left is the simulator \cite{okal2014towards} in RVIZ, and the right is our scene in Gazebo where the pedestrian data of the simulator is synchronized.}
	\label{fig:crowd}
	\vspace{-0.0cm}
\end{figure}

\djc{\textbf{Dense crowd experiment.}}
Our model is learned in a multi-agent scenario. Agents act as dynamic obstacles to each other, and all agents share the same policy. To show that our model is not only applicable to multi-agent scenarios with the same policy, but also to dense and complex crowd scenarios, we conduct dense crowd experiments. We use the pedestrian motion simulator \cite{okal2014towards} proposed by Billy et al. to generate pedestrian trajectories in RVIZ of ROS, use simple cubes instead of pedestrian models for testing in Gazebo, and transmit pedestrian coordinates from RVIZ to Gazebo in real-time, thus the cubes can simulate the movement of pedestrians. We deploy the robot model to test in three different scenarios (as shown in Fig.~\ref{fig:crowd}), randomly generate the initial position and the end position, compare the success rate and time of the five models, and each model is tested 100 times in each scenario. As shown in Table~\ref{dense crowd}, we can observe that it is similar to Table~\ref{example_stage}. \djc{The success rate of our $90\,^{\circ}$-FOV model is similar to the $180\,^{\circ}$-FOV CNN model, while the $90\,^{\circ}$-FOV CNN model performs poorly. Because pedestrians are too dense and have irregular movements, the success rate is generally low. Due to the deepening of the network, our policy has taken a relatively longer time.}

\begin{table*}[t]
	\setlength{\abovecaptionskip}{-0.2cm}
	\setlength{\belowcaptionskip}{-1.0cm}
	\caption{\scriptsize PERFORMANCE OF AGENTS WITH DIFFERENT MODELS IN 3D SCENES OF DIFFERENT COMPLEXITY.}
	\label{example_gazebo}
	\begin{center}
		\begin{tabular}{|c|c|c|c|c|c|c|c|c|}
			
			\hline
			\multicolumn{2}{|c|}{Model} & \emph{E}$_{Btm.Las.}$ & \emph{E}$_{Top.Las.}$ & \emph{E}$_{Dep.1-D.}$ & \emph{E}$_{Dep.Pol.}$ & \emph{E}$_{Dep.1-D.Sem.}$ & \tabincell{c}{\emph{E}$_{Dep.Pol.Sem.}$ \\ (proposed)} & \tabincell{c}{\emph{E}$_{Dep.Sem.Noi.}$  \\ (proposed)} \\
			\hline
			\multirow{2}{*}{Scene 7} 
			& Success rate & 23\% & 14\% & 0\% & 37\% & 6\%  & 82\% & \textbf{89\%}\\
			\cline{2-9}
			& Average time & 15.743 & 18.098 & - & 13.735 & - & 13.773 & \textbf{11.937} \\
			\hline
			\multirow{2}{*}{Scene 8} 
			& Success rate & \textbf{100\%} & 85\% & 77\% & 94\% & 85\% & \textbf{100\%} & \textbf{100\%}\\
			\cline{2-9}
			& Average time & \textbf{18.075} & 18.553 & 19.011 & 21.251 & 18.721 & 18.771 & 18.972 \\
			\hline
			\multirow{2}{*}{Scene 9} 
			& Success rate & 93\% & 87\% & 71\% & 94\% & 82\% & 94\% & \textbf{96\%} \\
			\cline{2-9}
			& Average time & 23.736 & \textbf{22.268} & 27.469 & 27.087 & 24.242 & 27.005 & 24.738 \\
			\hline
			\multirow{2}{*}{Scene 10} 
			& Success rate & 80\% & 70\% & 60\% & 70\% & 65\% & 80\% & \textbf{85\%} \\
			\cline{2-9}
			& Average time & \textbf{147.515} & 160.742 & 201.024 & 174.732 & 198.995 & 184.951 & 168.546 \\
			\hline
			
		\end{tabular}
	\end{center}
\end{table*}

\begin{figure*}[t]
	\centering
	\subfigure[cafe table]{
		\begin{minipage}[t]{0.19\linewidth}
			\centering
			\includegraphics[height=15mm, width=34mm]{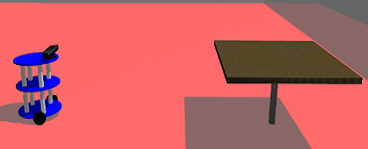}
		\end{minipage}%
	}%
	\subfigure[table]{
		\begin{minipage}[t]{0.19\linewidth}
			\centering
			\includegraphics[height=15mm, width=34mm]{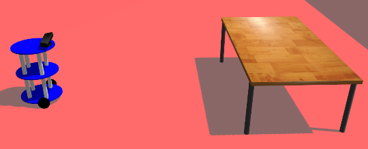}
		\end{minipage}%
	}%
	\subfigure[fire hydrant]{
		\begin{minipage}[t]{0.18\linewidth}
			\centering
			\includegraphics[height=15mm, width=34mm]{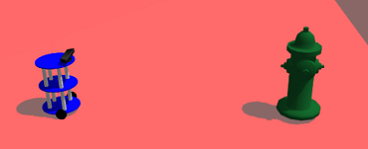}
		\end{minipage}
	}%
	\subfigure[construction cone]{
		\begin{minipage}[t]{0.18\linewidth}
			\centering
			\includegraphics[height=15mm, width=34mm]{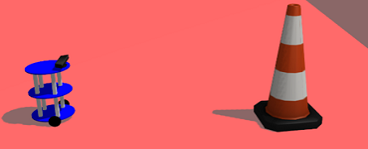}
		\end{minipage}
	}%
	\subfigure[cabinet]{
		\begin{minipage}[t]{0.18\linewidth}
			\centering
			\includegraphics[height=15mm, width=34mm]{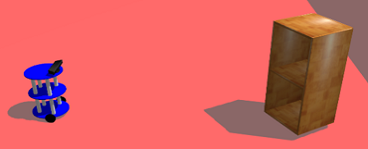}
		\end{minipage}
	}%
	\centering
	\caption{Test scenario of a single irregular obstacle deployed in Gazebo, including coffee table, table, fire hydrant, construction cone and cabinet.}
	\label{fig:single}

\end{figure*}

\begin{figure}[t]
	\centering
	\vspace{-0.2cm}
	\includegraphics [height=61.15mm,width=75mm ]{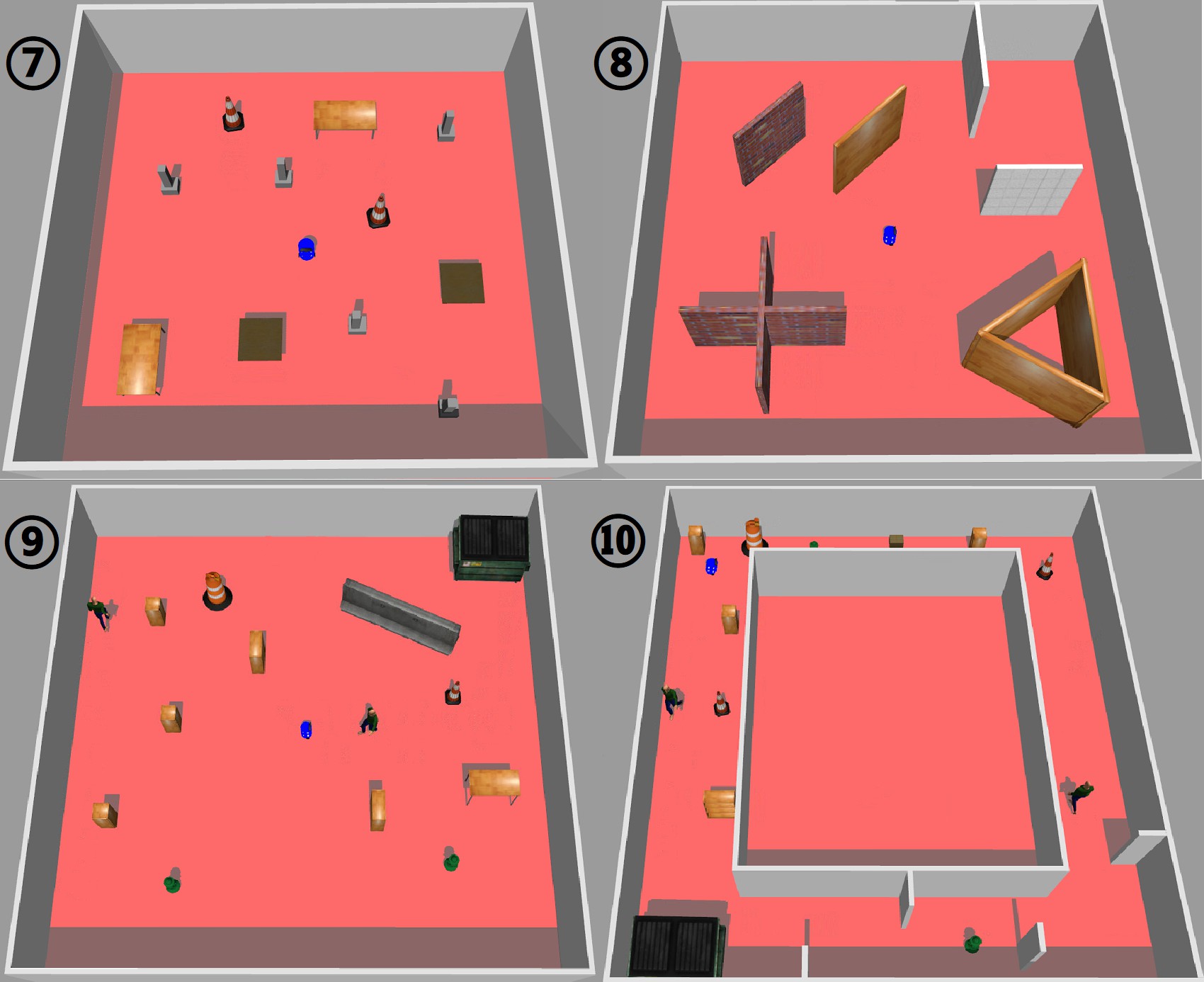}
	\caption{Different 3D test scenarios are used for our experiments. (7) A scene consisting of only irregular objects. (8) A simple static scene. (9) A more complex scene with pedestrians. (10) A complex corridor scene.}
	\label{fig:gazebo}
	\vspace{-0.2cm}
\end{figure}

\textbf{The entire monocular camera-based method.}
To demonstrate the effectiveness of our pseudo-laser measurement, we perform a visual simulation experiment. We build a 3D simulation environment based on Gazebo, as shown in Fig.\ref{fig:gazebo}. Scene 7 consists only of challenging irregular objects, while Scene 8-10 are complex corridor scenes with pedestrians. We build a simple two-wheel differential robot in Gazebo, deploying laser sensors and RGB-D cameras. The height of the robot is higher than the height of the table in the scene. In order to make the depth estimation and semantic segmentation models suitable for the Gazebo simulation environment, we collect scene-specific image data and pretrain the model for 6 hours so that it can be applied to virtual scenes. In order to improve the efficiency of image processing, we reduce the depth map by 5 times. The model runs on a computer with NVIDIA GTX 1080Ti GPU and an i7-7700 CPU, which can reach 10 frames per second.

We compare 7 models, lasers on the bottom of the robot (\emph{E}$_{Btm.Las.}$), laser on top of the robot (\emph{E}$_{Top.Las.}$), one-dimensional slice of depth map (\emph{E}$_{Dep.1-D.}$), dynamic local minimum pooling of depth map (\emph{E}$_{Dep.Pol.}$), one-dimensional slice of depth map combined with semantic mask (\emph{E}$_{Dep.1-D.Sem.}$), dynamic local minimum pooling of depth maps combined with semantic mask (\emph{E}$_{Dep.Pol.Sem.}$), and dynamic local minimum pooling of depth map and semantic mask with noise (\emph{E}$_{Dep.Sem.Noi.}$). We randomly generate a starting position and a goal position for a robot in each scene and test it 100 times. The corridor scene is set with 4 waypoints, and the robot moves circularly in the scene and tests for 20 times. The results are shown in Table~\ref{example_gazebo}.

\begin{table*}[ht]
	\setlength{\abovecaptionskip}{-0.2cm}
	\setlength{\belowcaptionskip}{-1.0cm}
	\caption{\scriptsize PERFORMANCE OF AGENTS WITH DIFFERENT MODELS FOR DIFFERENT IRREGULAR OBSTACLES.}
	\label{single obstacle}
	\begin{center}
		\begin{tabular}{|c|c|c|c|c|c|c|c|c|}
			
			\hline
			\multicolumn{2}{|c|}{Model} & \emph{E}$_{Btm.Las.}$ & \emph{E}$_{Top.Las.}$ & \emph{E}$_{Dep.1-D.}$ & \emph{E}$_{Dep.Pol.}$ & \emph{E}$_{Dep.1-D.Sem.}$ & \tabincell{c}{\emph{E}$_{Dep.Pol.Sem.}$ \\ (proposed)} & \tabincell{c}{\emph{E}$_{Dep.Sem.Noi.}$  \\ (proposed)} \\
			\hline 
			cafe\_table & Success rate & 0\% & 72\% & 24\% & 68\% & 18\%  & 90\% & \textbf{92\%}\\
			\hline
			table & Success rate & 4\% & 66\% & 18\% & 66\% & 22\%  & 88\% & \textbf{90\%}\\
			\hline
			fire\_hydrant & Success rate & 94\% & 92\% & 86\% & 80\% & 88\%  & 96\% & \textbf{98\%}\\
			\hline
			construction cone & Success rate & 96\% & 44\% & 52\% & 86\% & 38\%  & 92\% & \textbf{96\%}\\
			\hline
			cabinet & Success rate & 26\% & 34\% & 38\% & 64\% & 56\%  & 90\% & \textbf{94\%}\\
			\hline
			
		\end{tabular}
	\end{center}
\end{table*}

We observe that, in Scene 7 consisting only of irregular objects, almost all the comparison models fail to accomplish the task. Only our slicing method takes the shape of the obstacle into account and achieves the best performance. \djc{Since the model \emph{E}$_{Btm.Las.}$ considers the shape of the bottom of obstacles only, it reacts poorly towards obstacles such as desks.} Similarly, the model \emph{E}$_{Top.Las.}$ can only scan obstacle information at a certain height, which is not able to handle lower objects and cone objects. The model \emph{E}$_{Dep.1-D.}$ does not perform well for all objects since it falsely integrates the depth of the traversable region. The model \emph{E}$_{Dep.Pol.}$ shows some successful collision avoidance behaviors for some objects. However, the traversable region information is not considered, therefore the navigation success rate is still poor. Although the model \emph{E}$_{Dep.1-D.Sem.}$ considers the depth of the traversable region, it still only captures one-dimensional data, which is not effective. \djc{In contrast, the model \emph{E}$_{Dep.Pol.Sem.}$ considers both the depth of traversable region and the overall shape of the obstacle, which can achieve a better success rate of obstacle avoidance. We also test in more complex scenes, \emph{i.e.}, Scene 8-10, and the results show that our method has better robustness and generalization, and even better than laser sensors.} Based on this, we also introduce a model trained via our proposed data augmentation method \emph{E}$_{Dep.Sem.Noi.}$. During training, the data augmentation setting details are as below. For identifying junction boundaries in laser measurements, we set the threshold for adjacent values as 0.5, and the range of neighborhoods is set as 8. For non-boundaries, we add the Gaussian white noise to each value of the laser measurement for data augmentation. The scale of the noise is 0.07 times the value. \djc{Table~\ref{example_gazebo} shows that the addition of noise in data augmentation for training the model has the highest success rate. Without such data augmentation, the agent will behave hesitantly and it may rotate at a large angle due to the difference between the pseudo-laser estimated from the depth map and the actual laser sensing data in the virtual environment.}

	

\textbf{Single irregular object experiment.}
To further illustrate the effectiveness of our policy on obstacle avoidance for irregular objects, we tested 5 different irregular obstacles (cafe table, table, fire hydrant, construction cone, and cabinet) in the Gazebo environment (as shown in Fig.~\ref{fig:single}). For the robot to randomly generate an initial position in front of each obstacle and a target position after the obstacle, each obstacle is tested 50 times.
The results are shown in Table~\ref{single obstacle}. \djc{We can observe that all strategies have a high success rate for fire hydrant objects because it is similar to a cylindrical shape and is more regular than other objects. However, for the conical object and the table, the simple laser obstacle avoidance policy and the simple depth slicing policy are not ideal.} And our policy has a higher success rate of obstacle avoidance for all irregular objects, showing good generalization.

\begin{figure}[t]
	\setlength{\tabcolsep}{4pt}\small{
		\begin{tabular}{ccc}
			
			(A) & (B) & (C) \\
			\textbf{Scene with water} & \textbf{Scene with slope} & \textbf{Scene with clothes}  \\

			\includegraphics[height=18.75mm,width=25mm]{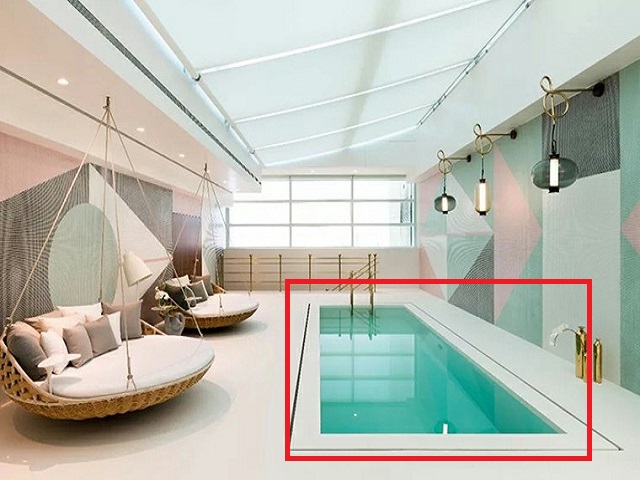}  &
			\includegraphics[height=18.75mm,width=25mm]{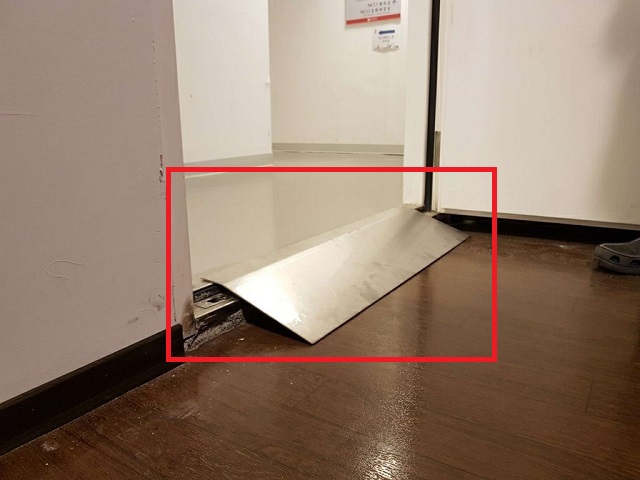}  &
			\includegraphics[height=18.75mm,width=25mm]{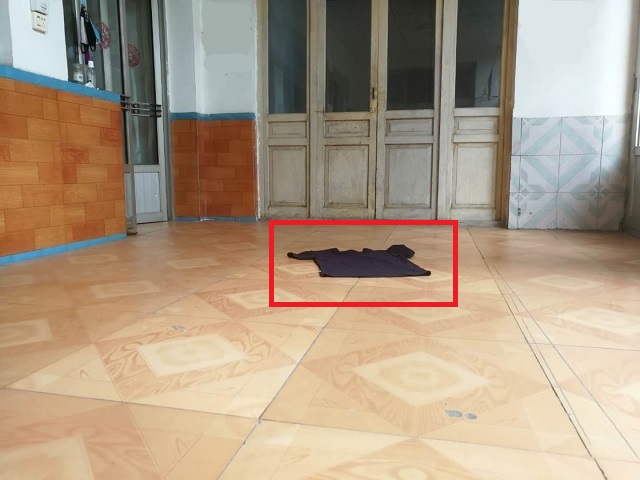} \\
			
			RGB image & RGB image & RGB image  \\
			
			\includegraphics[height=18.75mm,width=25mm]{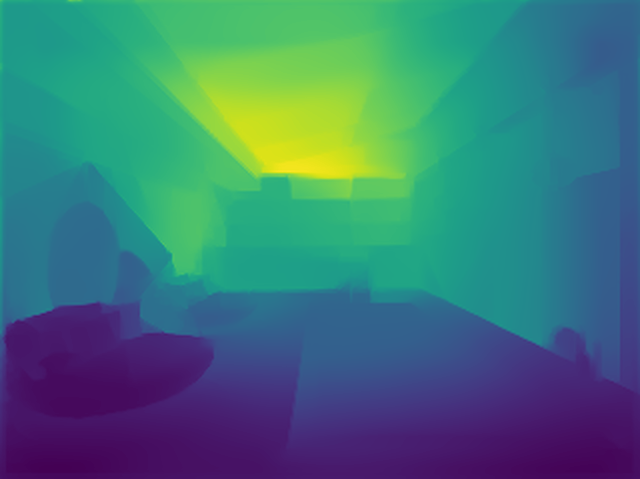}  &
			\includegraphics[height=18.75mm,width=25mm]{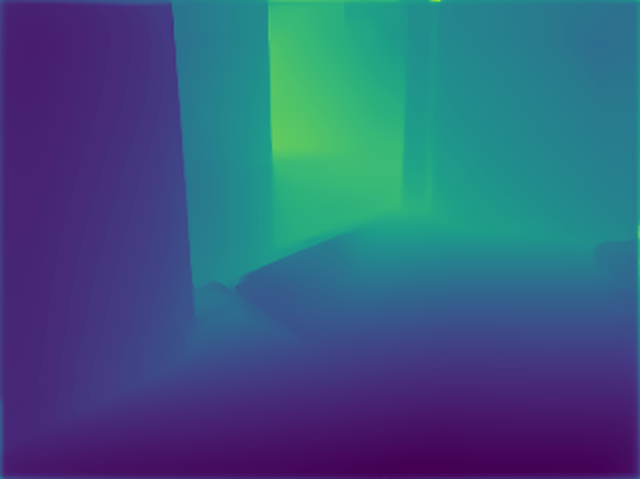}  &
			\includegraphics[height=18.75mm,width=25mm]{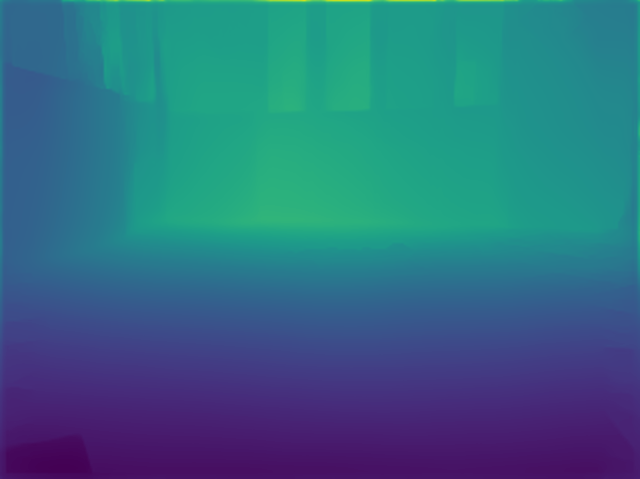} \\
			
			Depth image & Depth image & Depth image  \\

			\includegraphics[height=18.75mm,width=25mm]{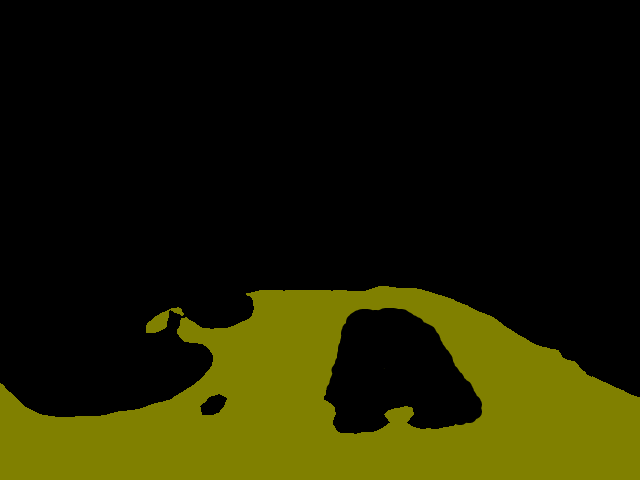}  &
			\includegraphics[height=18.75mm,width=25mm]{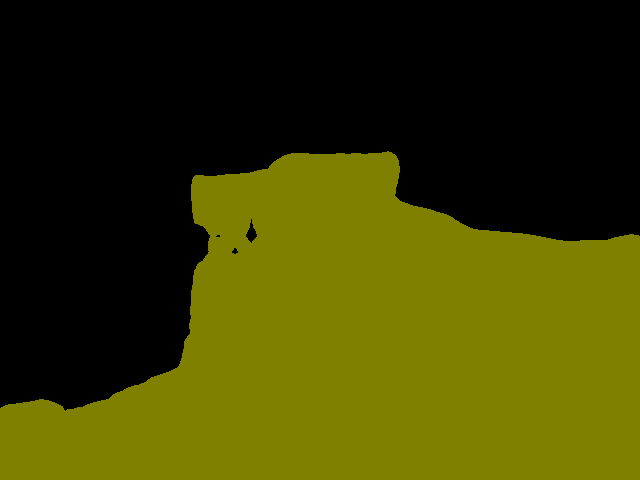}  &
			\includegraphics[height=18.75mm,width=25mm]{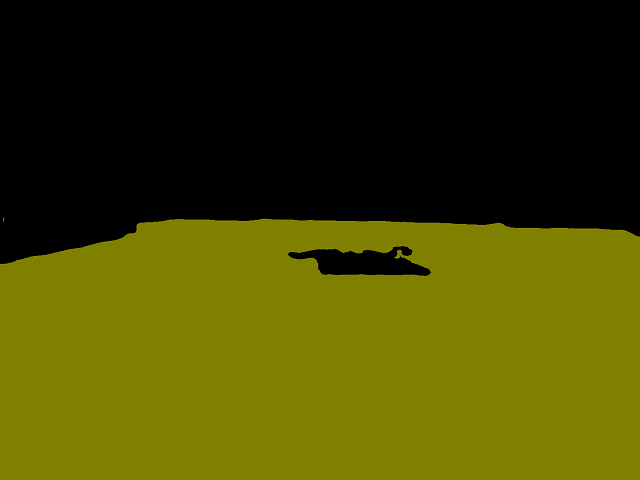} \\
			
			Semantic image & Semantic image & Semantic image  \\
			
			\includegraphics[height=18.75mm,width=25mm]{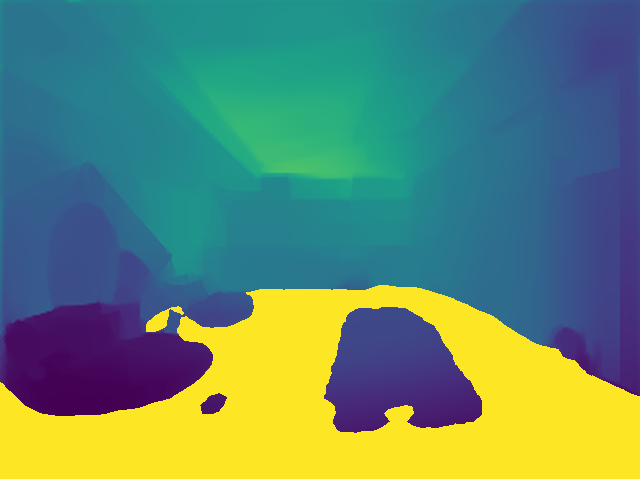}  &
			\includegraphics[height=18.75mm,width=25mm]{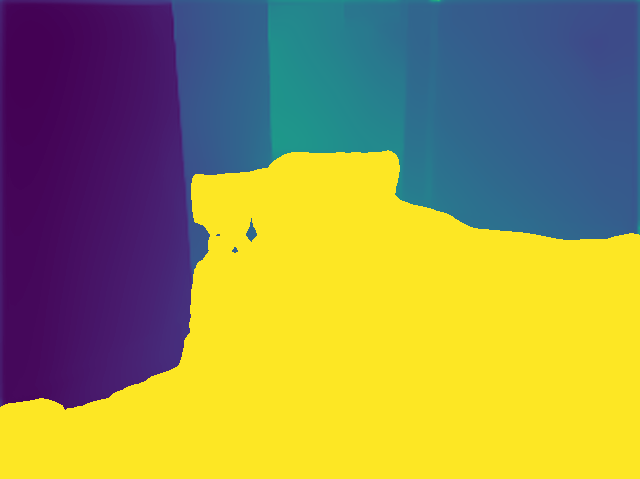}  &
			\includegraphics[height=18.75mm,width=25mm]{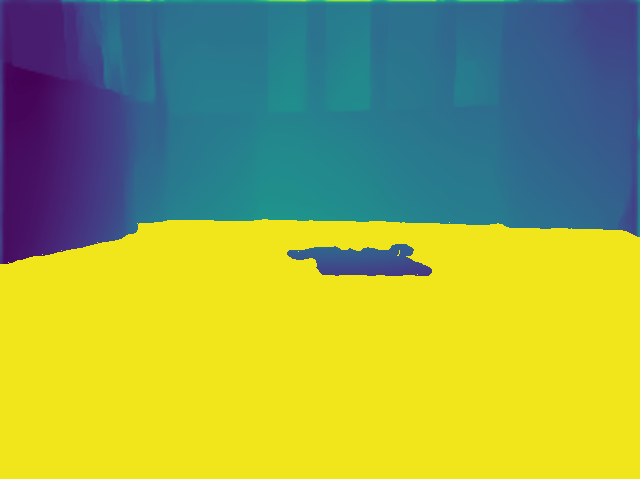} \\
			
			\footnotesize Semantic-depth image & \footnotesize Semantic-depth image & \footnotesize Semantic-depth image  \\
			
			\includegraphics[height=18.75mm,width=25mm]{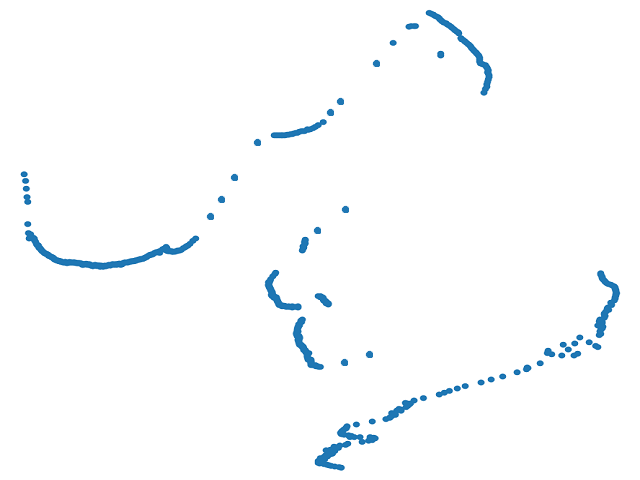}  &
			\includegraphics[height=18.75mm,width=25mm]{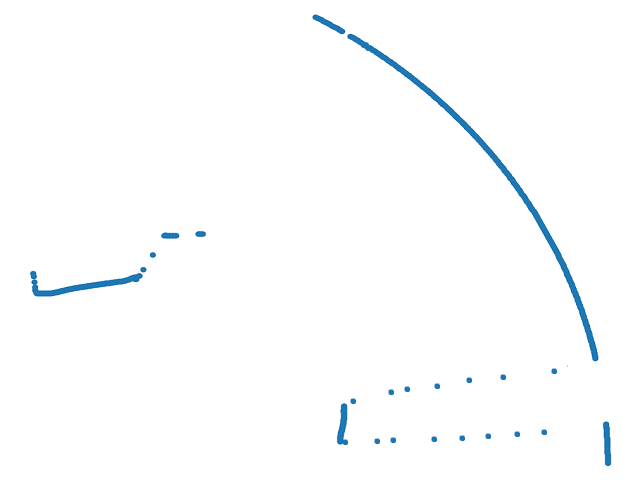}  &
			\includegraphics[height=18.75mm,width=25mm]{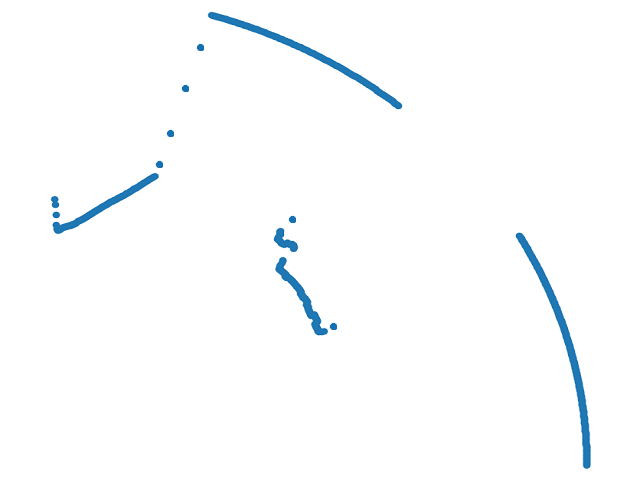} \\
			
			\tabincell{c}{\footnotesize Pseudo-laser \\\footnotesize  measurement}  &  \tabincell{c}{\footnotesize Pseudo-laser \\\footnotesize  measurement}  &  \tabincell{c}{\footnotesize Pseudo-laser \\\footnotesize  measurement}   \\
			
	\end{tabular}}
	
	\caption{Visualization of the process of acquiring pseudo-laser measurement using deep slicing in some complex ground scenes. Each column represents a scene (\emph{i.e.}, a scene with a water surface, a scene with a slope and a scene with clothes on the floor), and each row represents the image processing procedure (including RGB image, depth image, semantic image, semantic-depth image and pseudo-laser measurement).}
	
	\label{fig:semantic}
	\vspace{-0.6cm}
\end{figure}

\textbf{Complex ground scenes experiment.}
In order to further explain the role of our fusion of semantic information, as shown in Fig.~\ref{fig:semantic}, we visualize the process of acquiring pseudo-laser measurement in three complex scenarios. Taking Fig.~\ref{fig:semantic}(A) as an example, we obtain a depth map from an RGB image and a binary classification semantic mask with only the traversable region and background. After fusing into a semantic-depth map, we use the depth slicing method we proposed to obtain the final pseudo-laser measurement. \djc{The traditional method obtains laser measurement from the depth map by horizontal slice mapping of the point cloud. Since semantic information is not considered, the water surface is also cut off from the ground.} However, in reality, the water surface is not the safe passable region of the robot. \djc{Therefore we use semantic information to view the water as an unpassable region. The depth information of the water surface is also taken into account during the slicing process, thus providing the robot with the safest pseudo-laser measurement.} In Fig.~\ref{fig:semantic}(B), there is a highly misaligned slope. For a simple horizontal slice, it is difficult to set a threshold, resulting in ground depth affecting navigation. Through semantic information, we can obtain the traversable region of robots, so as to generate the most efficient and reasonable pseudo-laser measurement. There is a piece of clothing on the ground in Fig.~\ref{fig:semantic}(C). In fact, the clothing cannot be crushed and should be regarded as an obstacle. We can obtain the pseudo-laser measurement considering the depth of the clothing. \djc{Therefore, the method of obtaining the semantic mask of the traversable region through the semantic information and then performing depth slicing can enable the robot to effectively deal with scenes with the complex ground.}

\begin{figure}[t]
	\setlength{\tabcolsep}{1pt}\small{
		\begin{tabular}{cccc}
			
			(A) & (B) & (C) & (D) \\
			\textbf{Chair} & \textbf{Table} & \textbf{Clothes} & \textbf{Glass}  \\

			\includegraphics[width=0.24\linewidth]{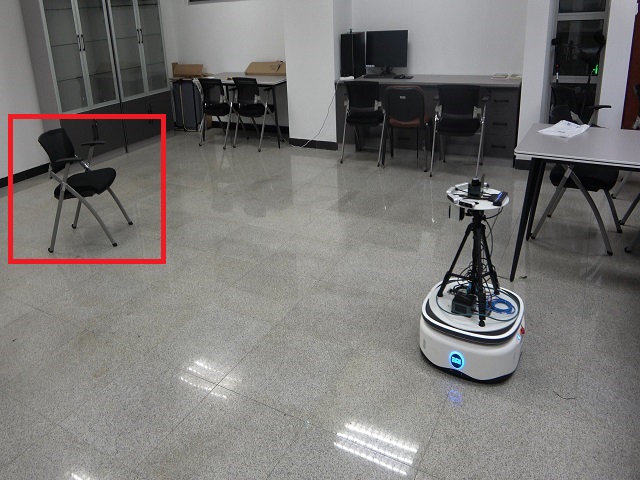}  &
			\includegraphics[width=0.24\linewidth]{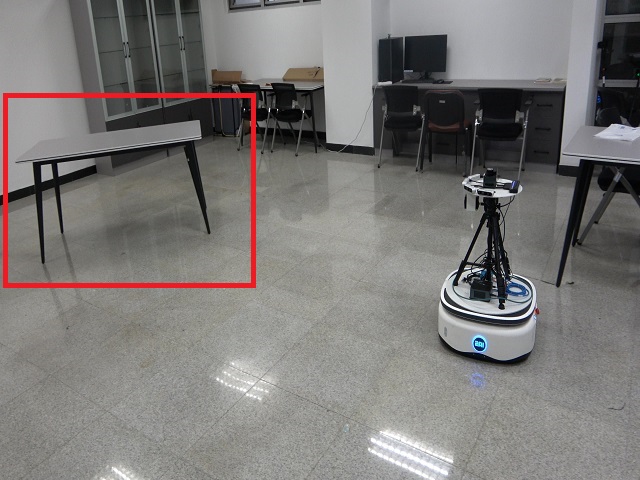}  &
			\includegraphics[width=0.24\linewidth]{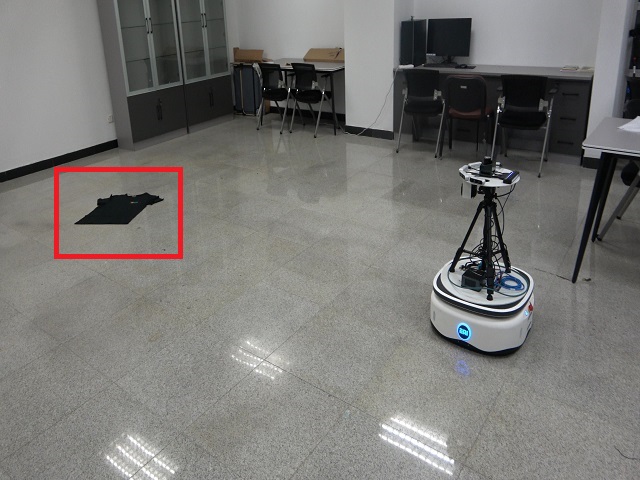} &
			\includegraphics[width=0.24\linewidth]{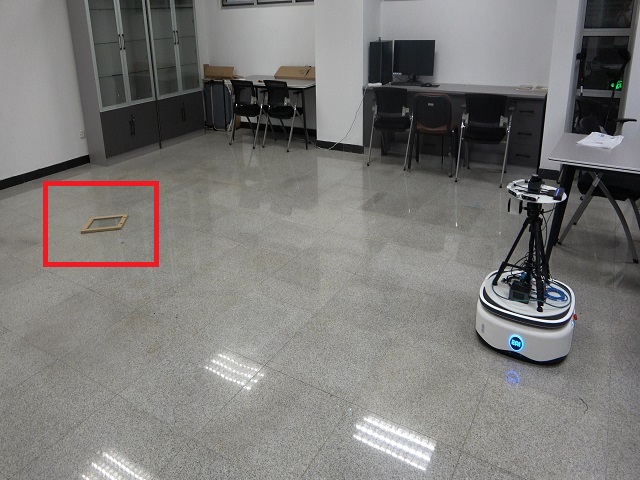} \\
			
			Observation & Observation & Observation & Observation \\
			
			\includegraphics[width=0.24\linewidth]{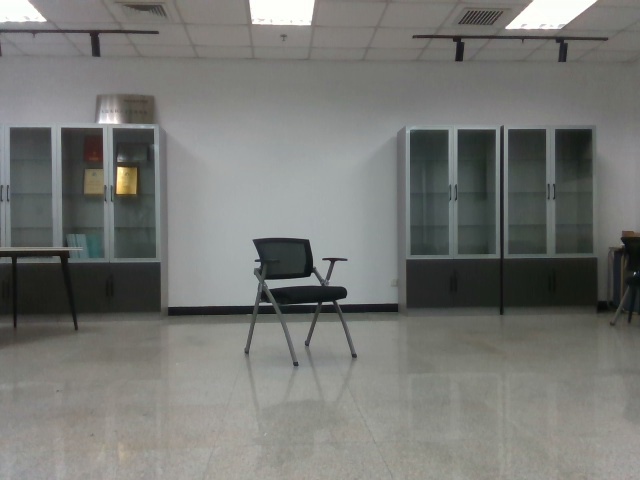}  &
			\includegraphics[width=0.24\linewidth]{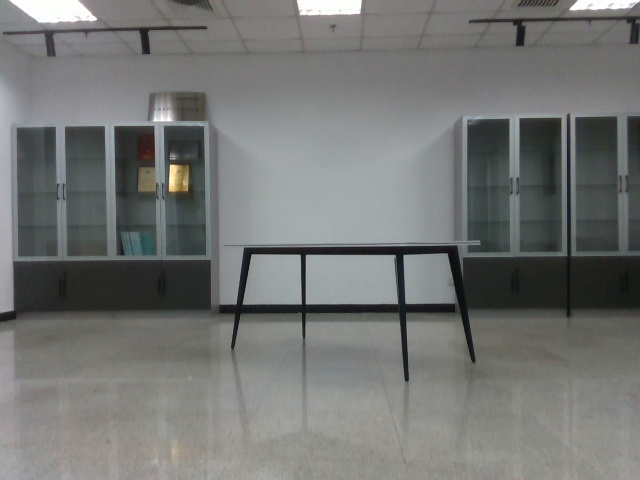}  &
			\includegraphics[width=0.24\linewidth]{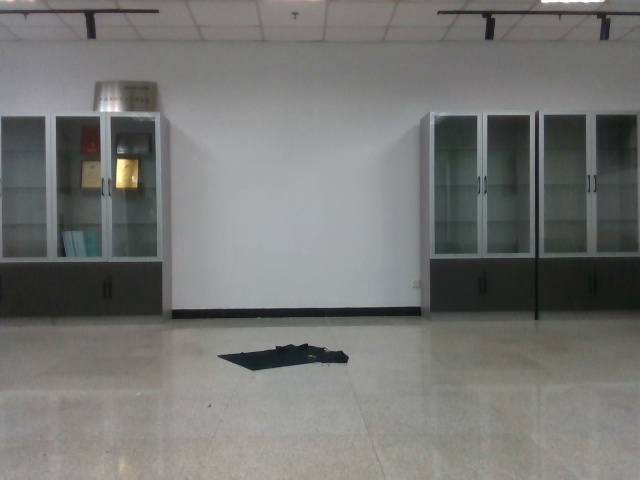} &
			\includegraphics[width=0.24\linewidth]{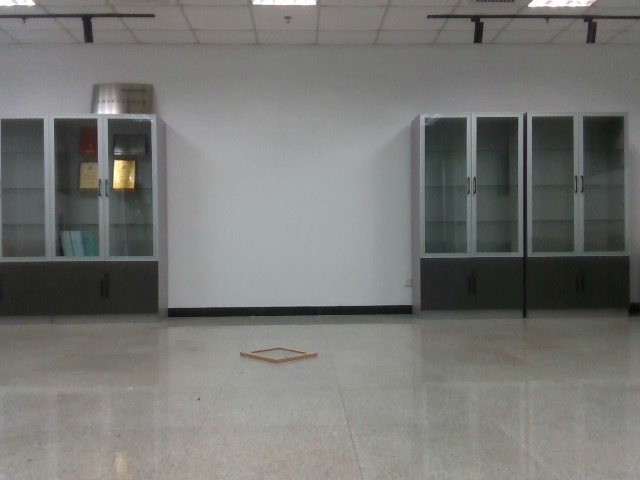} \\
			
			RGB image & RGB image & RGB image  & RGB image\\

			\includegraphics[width=0.24\linewidth]{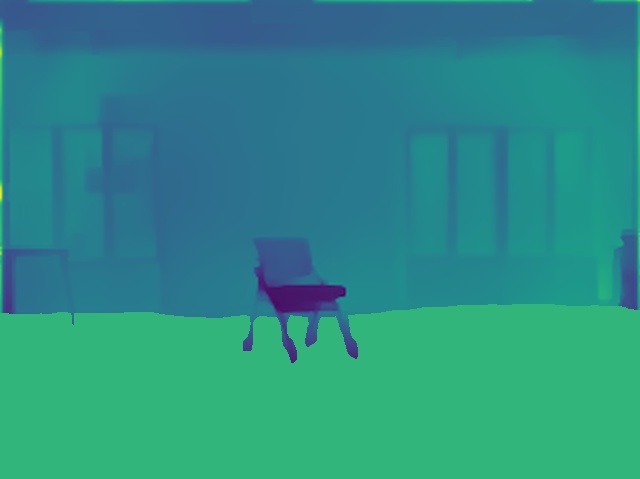}  &
			\includegraphics[width=0.24\linewidth]{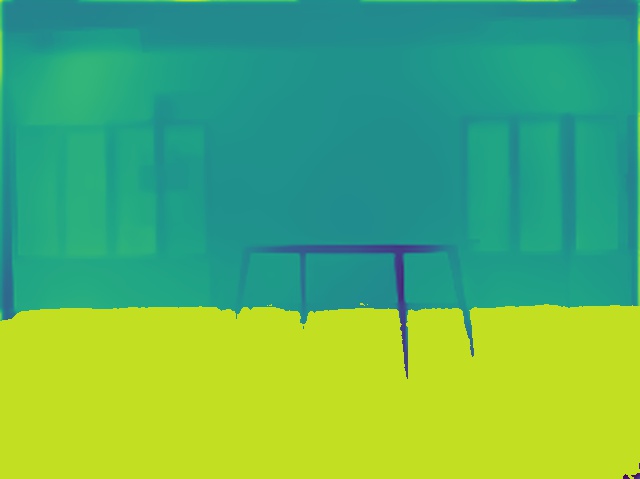}  &
			\includegraphics[width=0.24\linewidth]{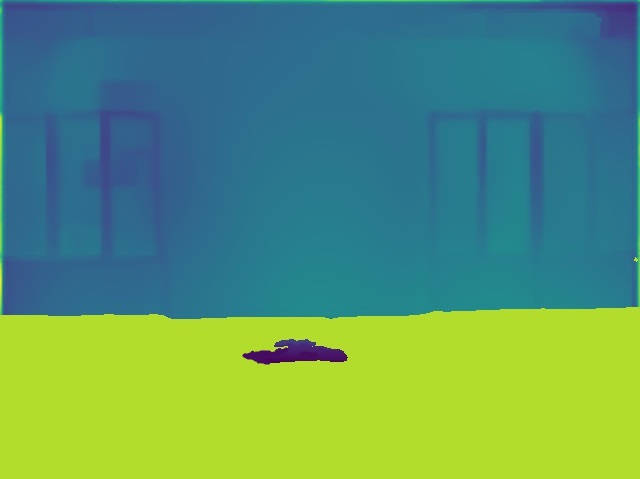} &
			\includegraphics[width=0.245\linewidth]{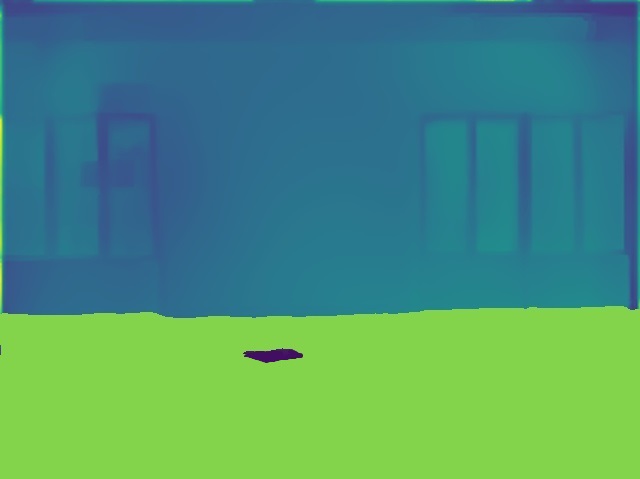} \\
			
			\tabincell{c}{\footnotesize Semantic-depth \\ \footnotesize image} & \tabincell{c}{\footnotesize Semantic-depth \\ \footnotesize image} & \tabincell{c}{\footnotesize Semantic-depth \\ \footnotesize image} & \tabincell{c}{\footnotesize Semantic-depth \\ \footnotesize image} \\

			\includegraphics[width=0.24\linewidth]{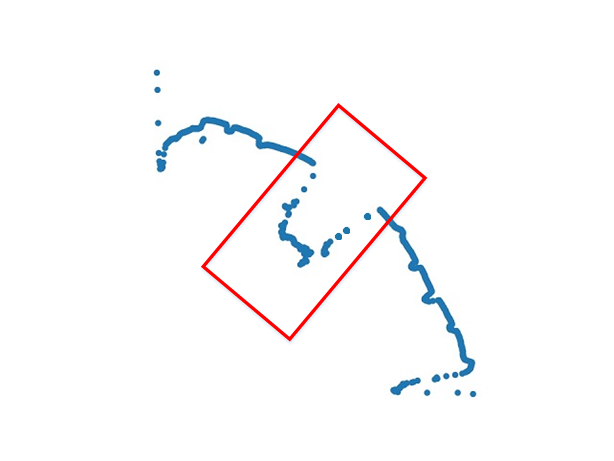}  &
			\includegraphics[width=0.24\linewidth]{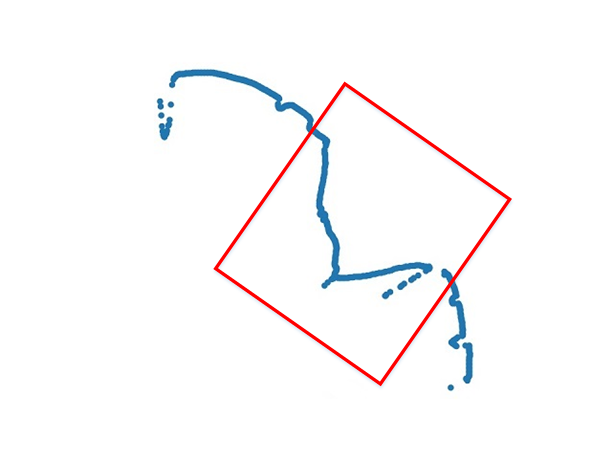}  &
			\includegraphics[width=0.24\linewidth]{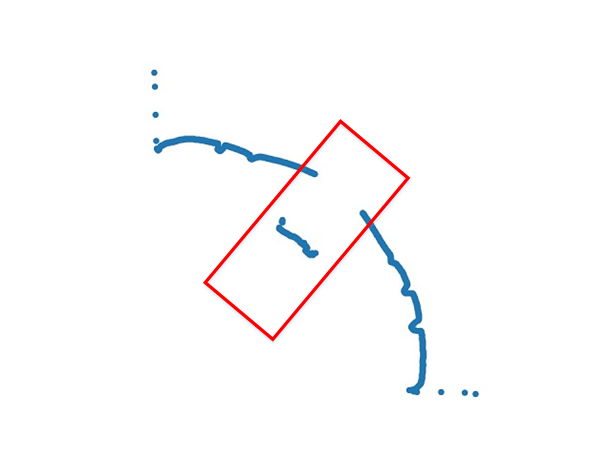} &
			\includegraphics[width=0.24\linewidth]{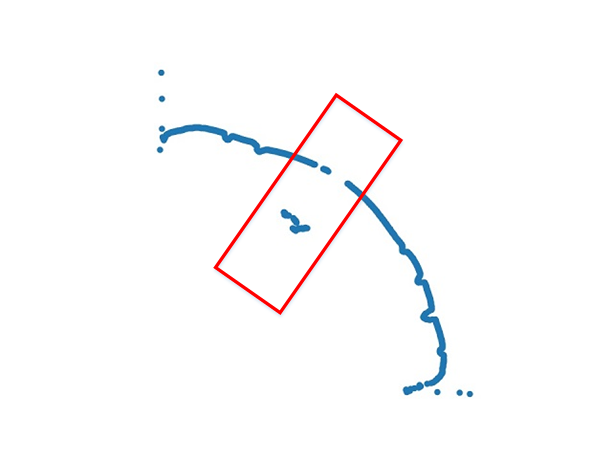} \\
			
			\tabincell{c}{\footnotesize Pseudo-laser \\\footnotesize  measurement}  &  \tabincell{c}{\footnotesize Pseudo-laser \\\footnotesize  measurement}  &  \tabincell{c}{\footnotesize Pseudo-laser \\\footnotesize  measurement}  &  \tabincell{c}{\footnotesize Pseudo-laser \\\footnotesize  measurement}  \\
			
			\includegraphics[width=0.24\linewidth]{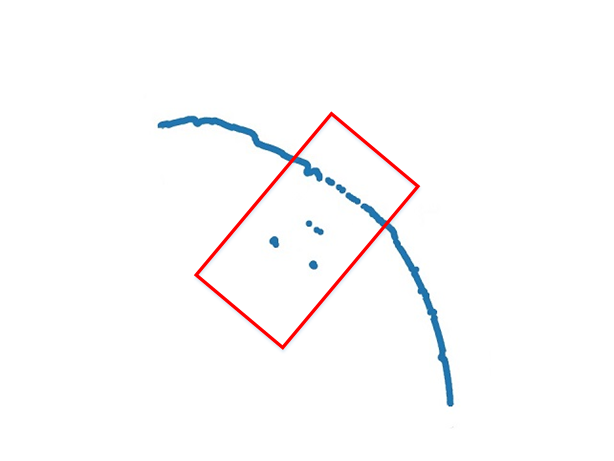}  &
			\includegraphics[width=0.24\linewidth]{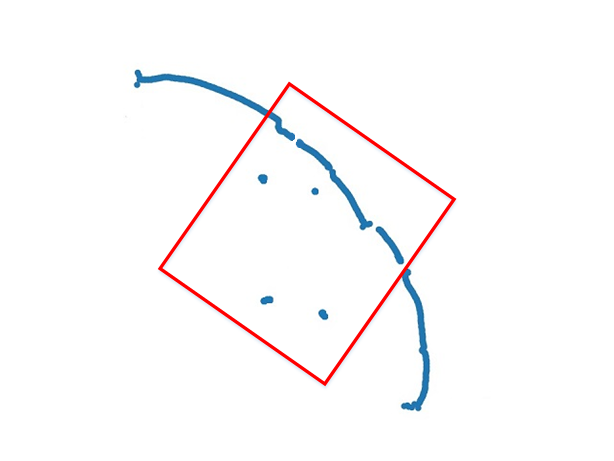}  &
			\includegraphics[width=0.24\linewidth]{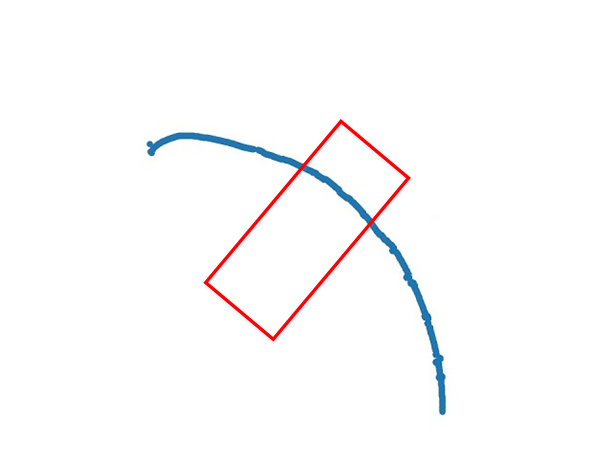} &
			\includegraphics[width=0.24\linewidth]{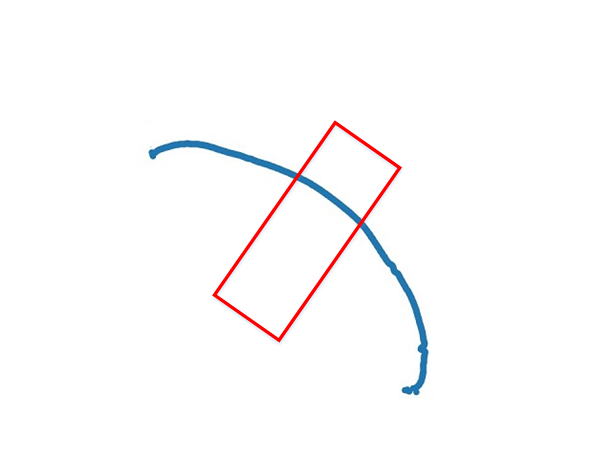} \\
			
			\tabincell{c}{\footnotesize laser device  \\  \footnotesize observation}  & \tabincell{c}{\footnotesize laser device  \\  \footnotesize observation} & \tabincell{c}{\footnotesize laser device  \\  \footnotesize observation} & \tabincell{c}{\footnotesize laser device  \\  \footnotesize observation} \\
			
	\end{tabular}}
	
	\caption{\djc{Visualize the process of acquiring pseudo laser to face complex obstacles in real-world scenes. Each column represents a complex obstacle (\emph{i.e.}, a chair, a table, clothes and glass), and each row represents the image processing procedure (including observation, RGB image, semantic-depth image, pseudo-laser measurement and bottom-laser device observation data).}}
	
	\label{fig:real_semantic}
	\vspace{-0.2cm}
\end{figure}

\textbf{Visualization experiment of complex obstacles in real-world scenes.} 
We deploy our method in a real-world scene, and visualize the generation process of pseudo-laser measurement when faced with complex objects, as shown in Fig.~\ref{fig:real_semantic}. Compared with laser device observation data, our approach with semantic information can effectively segment the traversable region, extract the edge contours of irregular objects, and identify ‘special obstacles’ in complex ground scenes (as shown in the red boxes in Fig.~\ref{fig:real_semantic}). Thereby our framework can perform efficient DRL to obtain reasonable actions through the generated pseudo-laser measurement. See the supplementary video for more details.

\begin{figure}[t]
	\centering
	
	\includegraphics [height=54mm,width=83mm ]{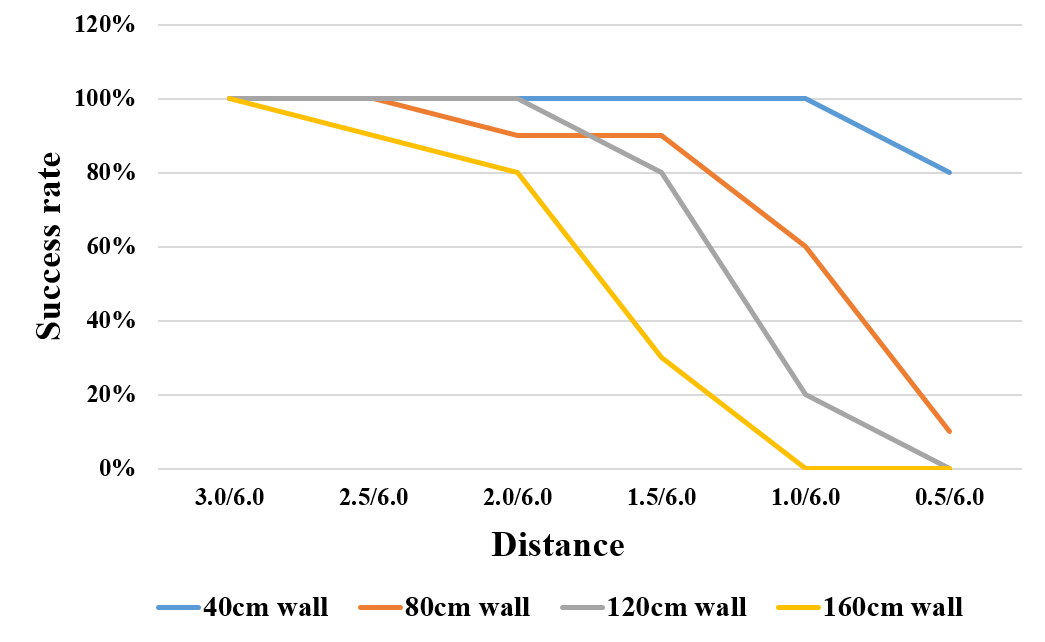}
	\caption{\djc{Experimental results of the limitations of the monocular-visual obstacle avoidance framework for obstacles of different sizes with different initial distances.}}
	\label{fig:limitation}
	\vspace{-0.2cm}
\end{figure} 

\begin{figure*}[t]
	\centering
	
	\includegraphics [height=73.82mm,width=182mm ]{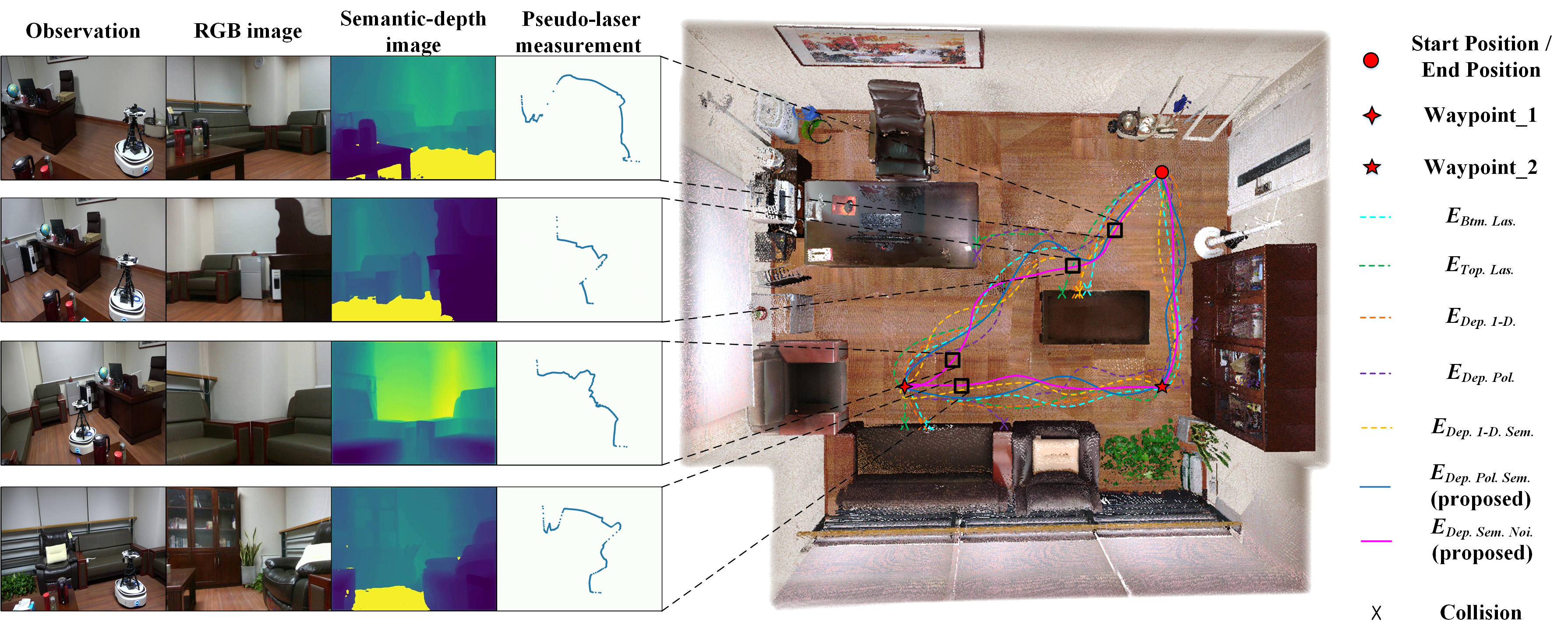}
	\caption{\djc{The trajectories generated by different models during navigating an agent in an office scene with two waypoints and visualization of the results at key locations.}}
	\label{fig:office}
	\vspace{-0.2cm}
\end{figure*}

\textbf{Monocular-visual limitation experiment.}
Visual obstacle avoidance will be affected by the current camera perspective. Although our policy will avoid obstacles in advance when facing obstacles, in order to study the limitations of our method, we manually initialize the location of obstacles to test the navigation success rate.
We find that navigation failed when obstacles occupied most of the FOV, thus different obstacles have different navigation safety distances. We use walls of different sizes as test obstacles, and each wall tests multiple different distances.

\begin{itemize}
	
\item \textbf{Distance} - The quotient of the distance between the robot and the obstacle and the maximum distance taken by the virtual camera (6.0m).

\end{itemize}

The result is shown in Fig.~\ref{fig:limitation}. The navigation success rate is lower when the initial position is closer to the obstacle. \djc{Obviously, the greater the space occupied by obstacles, the greater the distance for policy failure. When the obstacle occupies 65\% of the maximum angle of view ($90\,^{\circ}$-FOV), our policy starts to perform poorly, and the success rate decreases. When it occupies 80\%, our policy will fail, showing a hover or even collision.}

\begin{figure}[t]
	\centering
	
	\includegraphics [height=32.54mm,width=88mm ]{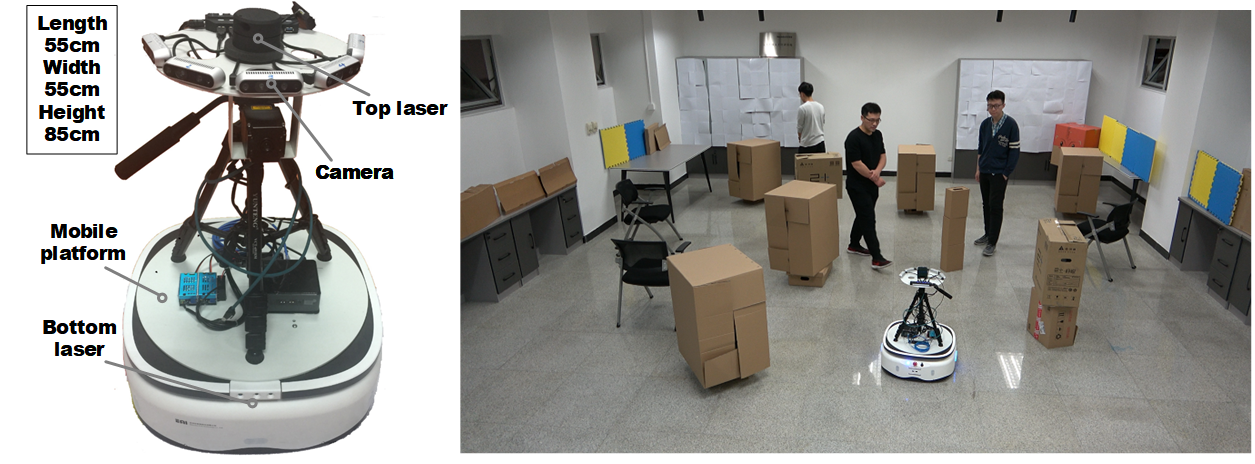}
	\caption{Illustration of the mobile robot platform used in our hardware experiments. It has 5 Intel Realsense D435 depth cameras, but we only use RGB images from \textbf{the middle one} of them. Two lasers at the top and bottom are used for testing.}
	\label{fig:robot}
	\vspace{-0.2cm}
\end{figure}

\begin{figure}[t]
	\centering
	
	\includegraphics [height=48mm,width=64mm ]{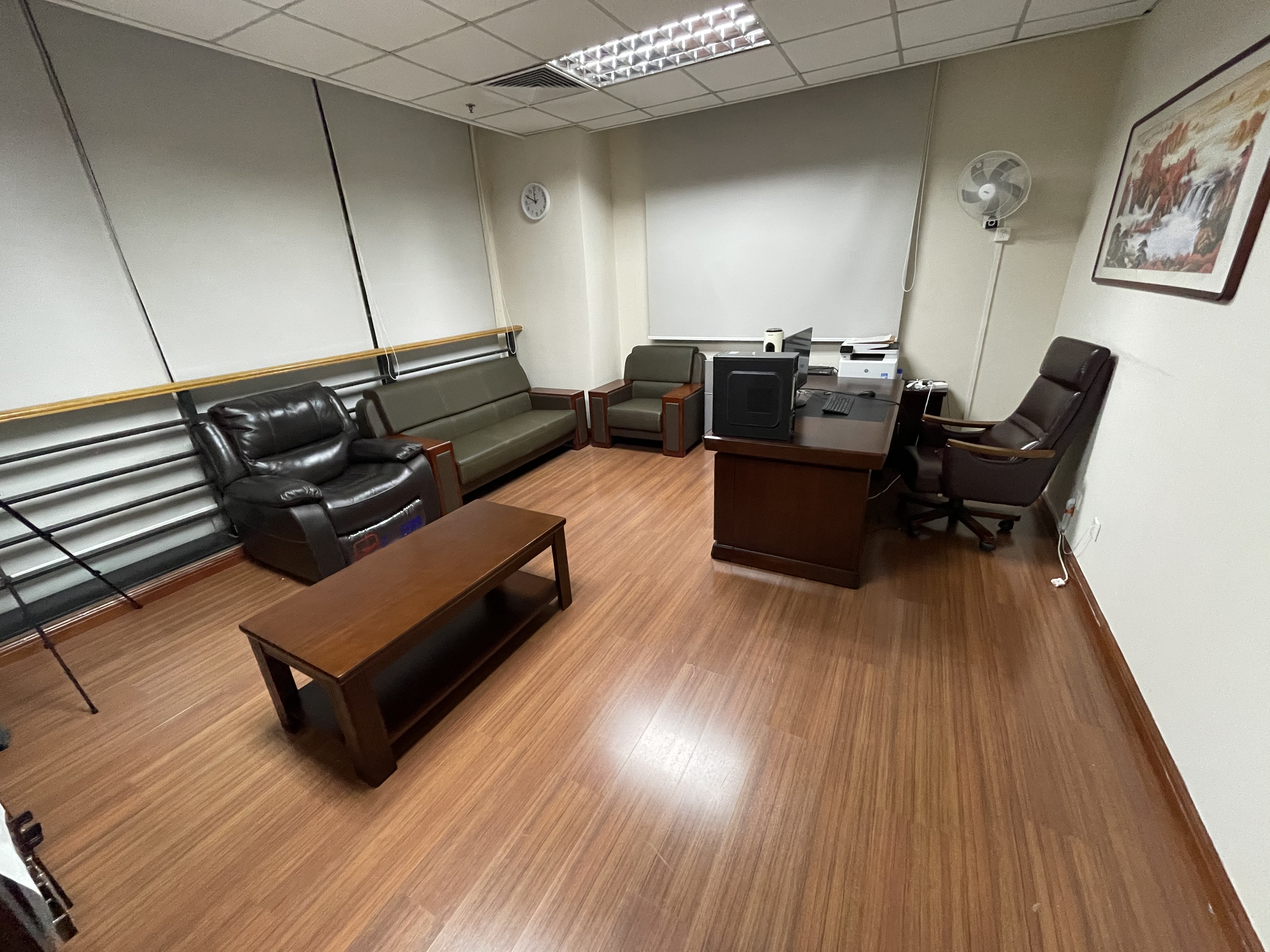}
	\caption{\djc{Our real test scene, a daily office.}}
	\label{fig:office-raw}
	\vspace{-0.2cm}
\end{figure}

\subsection{Hardware experiment}

In order to see the overall performance of our method in the real world, we deploy a ROS-based mobile robot as shown in Fig.~\ref{fig:robot} in a real scene constructed of cardboard boxes, tables, and chairs. The robot transmits the acquired images through a router to a local host with an Nvidia GTX 1080Ti GPU. Then the host performs image processing action prediction locally and sends speed information to the robot. The image processing can reach 10 frames per second in real-time. Our method can safely navigate and avoid obstacles in unseen scenes, and shows good robustness and generalization to static complex objects and challenging dynamic pedestrians.

\djc{In addition, we test the effect of different models in a daily office scene as shown in Fig.~\ref{fig:office-raw}, and visualized the trajectories of different models and the processing results of our method at key locations. As shown in Fig.~\ref{fig:office}, we visualize the trajectories of 7 models in the overhead point cloud of the office scene, and all 5 models collided except for our proposed method. Among them, the three models of \emph{E}$_{Btm.Las.}$, \emph{E}$_{Top.Las.}$ and \emph{E}$_{Dep.1-D.}$ are difficult to perceive the sofa and the shorter table, and they all collide. \emph{E}$_{Dep.Pol.}$ has difficulty in distinguishing the useless traversable region depth and mistakes the table as a path thus colliding with the large table. In our proposed method, \emph{E}$_{Dep.Pol.Sem.}$ does not collide but has a more tortuous route and behaves hesitantly, especially at locations with large depth variations. \emph{E}$_{Dep.Sem.Noi.}$ shows the best results, and we visualize the pseudo laser measurement generation process of \emph{E}$_{Dep.Sem.Noi.}$ at key locations where other models have a high chance of collision. The results show more intuitively that our proposed method can effectively deal with complex scenes.} See the supplementary video for more details.

\begin{figure}[t]
	\setlength{\tabcolsep}{5pt}\small{
		\begin{tabular}{cc}

			\includegraphics[height=30mm,width=40mm]{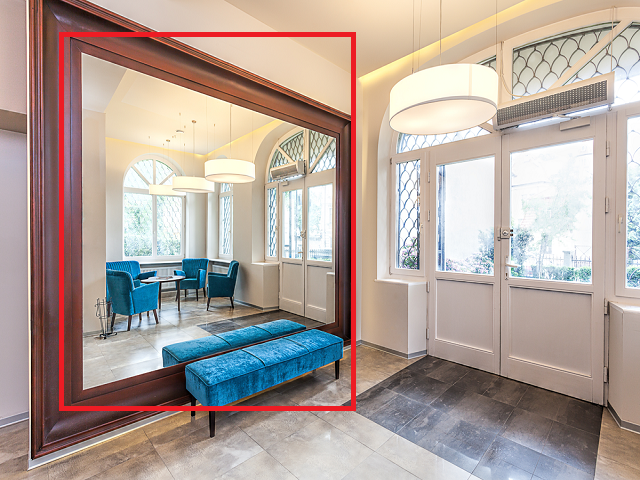}  &
			\includegraphics[height=30mm,width=40mm]{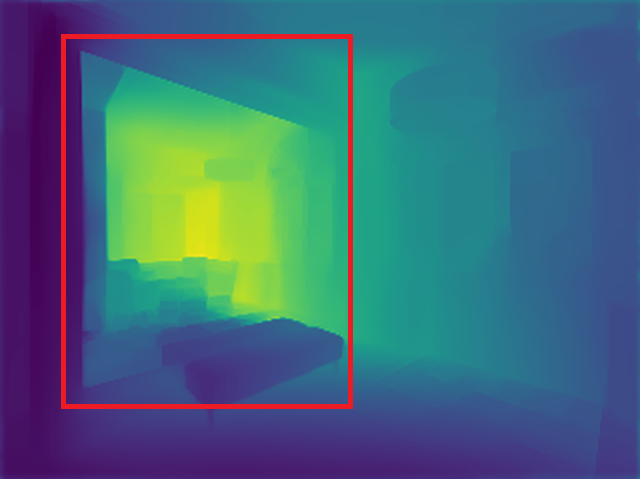}  \\
			
			(a) RGB image & (b) Depth image

	\end{tabular}}
	
	\caption{\djc{Failure cases. Our method may fail in a scene with mirrors because the depth estimation does not perform well for the depth of the mirror.}}
	
	\label{fig:failure}
\end{figure}

\section{Conclusion and Future work}

In this paper, we propose a framework that uses a low-cost monocular RGB camera to accomplish obstacle avoidance for mobile robots. Against the complex obstacles in challenging environments, we design the pseudo-laser, which fuses distance and semantic information and exhibits good collision avoidance performance in unseen complex scenes. The FEG module we proposed can reasonably weight the observed data, thus the mobile robot can capture the surrounding environment well and achieve efficient collision avoidance. In addition, we use data augmentation to improve the performance of the policy in the real-world and introduce an LSTM module to solve the problem with the limited FOV. However, our method has limitations. Due to the poor performance of the depth estimation method for the mirror, our method is invalid in some indoor scenes with mirrors~\cite{yang2019my}, as shown in Fig.~\ref{fig:failure}. \djc{In future work, we will consider more complex concepts, such as robot positioning via pure vision \cite{hirose2019deep}, visual servoing \cite{hutchinson1996tutorial} and the application of hierarchical reinforcement learning \cite{li2020hrl4in} in obstacle avoidance tasks.}

\section*{ACKNOWLEDGMENT}

This work was supported in part by National Key Research and Development Program of China (2021ZD0112400), the National Natural Science Foundation of China under Grant  61972067/U21A20491/U1908214, and the Innovation Technology Funding of Dalian (2020JJ26GX036).

	





\ifCLASSOPTIONcaptionsoff
  \newpage
\fi



%

\addtolength{\textheight}{0cm}   





\bibliographystyle{IEEEtranS} 

\bibliography{IEEEexample}

\begin{thebibliography}{10}
\providecommand{\url}[1]{#1}
\csname url@rmstyle\endcsname
\providecommand{\newblock}{\relax}
\providecommand{\bibinfo}[2]{#2}
\providecommand\BIBentrySTDinterwordspacing{\spaceskip=0pt\relax}
\providecommand\BIBentryALTinterwordstretchfactor{4}
\providecommand\BIBentryALTinterwordspacing{\spaceskip=\fontdimen2\font plus
\BIBentryALTinterwordstretchfactor\fontdimen3\font minus
  \fontdimen4\font\relax}
\providecommand\BIBforeignlanguage[2]{{%
\expandafter\ifx\csname l@#1\endcsname\relax
\typeout{** WARNING: IEEEtran.bst: No hyphenation pattern has been}%
\typeout{** loaded for the language `#1'. Using the pattern for}%
\typeout{** the default language instead.}%
\else
\language=\csname l@#1\endcsname
\fi
#2}}

\bibitem{arokiasami2016interoperable}
W.~A. Arokiasami, P.~Vadakkepat, K.~C. Tan, and D.~Srinivasan, ``Interoperable
  multi-agent framework for unmanned aerial/ground vehicles: towards robot
  autonomy,'' \emph{Compl. Intel. Syst.}, vol.~2, no.~1, pp. 45--59, 2016.

\bibitem{bharadhwaj2019data}
H.~Bharadhwaj, Z.~Wang, Y.~Bengio, and L.~Paull, ``A data-efficient framework
  for training and sim-to-real transfer of navigation policies,'' in
  \emph{Proc. IEEE Int. Conf. Robot. Autom.}, 2019, pp. 782--788.

\bibitem{bills2011autonomous}
C.~Bills, J.~Chen, and A.~Saxena, ``Autonomous mav flight in indoor
  environments using single image perspective cues,'' in \emph{Proc. IEEE Int.
  Conf. Robot. Autom.}, 2011, pp. 5776--5783.

\bibitem{chen2019learning}
G.~Chen, H.~Yu, W.~Dong, X.~Sheng, X.~Zhu, and H.~Ding, ``Learning to navigate
  from simulation via spatial and semantic information synthesis,''
  \emph{arXiv:1910.05758}, 2019.

\bibitem{chen2017socially}
Y.~F. Chen, M.~Everett, M.~Liu, and J.~P. How, ``Socially aware motion planning
  with deep reinforcement learning,'' in \emph{Proc. IEEE/RSJ Int. Conf.
  Intell. Robots Syst.}, 2017, pp. 1343--1350.

\bibitem{choi2019deep}
J.~Choi, K.~Park, M.~Kim, and S.~Seok, ``Deep reinforcement learning of
  navigation in a complex and crowded environment with a limited field of
  view,'' in \emph{Proc. IEEE Int. Conf. Robot. Autom.}, 2019, pp. 5993--6000.

\bibitem{cosio2004autonomous}
F.~A. Cos{\'\i}o and M.~P. Casta{\~n}eda, ``Autonomous robot navigation using
  adaptive potential fields,'' \emph{Mathematical and Computer Modelling},
  vol.~40, no. 9-10, pp. 1141--1156, 2004.

\bibitem{fan2018crowdmove}
T.~Fan, X.~Cheng, J.~Pan, D.~Manocha, and R.~Yang, ``Crowdmove: Autonomous
  mapless navigation in crowded scenarios,'' \emph{arXiv:1807.07870}, 2018.

\bibitem{fan2020distributed}
T.~Fan, P.~Long, W.~Liu, and J.~Pan, ``Distributed multi-robot collision
  avoidance via deep reinforcement learning for navigation in complex
  scenarios,'' \emph{The International Journal of Robotics Research}, vol.~39,
  no.~7, pp. 856--892, 2020.

\bibitem{faust2018prm}
A.~Faust, K.~Oslund, O.~Ramirez, A.~Francis, L.~Tapia, M.~Fiser, and
  J.~Davidson, ``Prm-rl: Long-range robotic navigation tasks by combining
  reinforcement learning and sampling-based planning,'' in \emph{Proc. IEEE
  Int. Conf. Robot. Autom.}, 2018, pp. 5113--5120.

\bibitem{fox1997dynamic}
D.~Fox, W.~Burgard, and S.~Thrun, ``The dynamic window approach to collision
  avoidance,'' \emph{IEEE Robotics \& Automation Magazine}, vol.~4, no.~1, pp.
  23--33, 1997.

\bibitem{fragkiadaki2015learning}
K.~Fragkiadaki, P.~Agrawal, S.~Levine, and J.~Malik, ``Learning visual
  predictive models of physics for playing billiards,''
  \emph{arXiv:1511.07404}, 2015.

\bibitem{gao2021vision}
L.~Gao, J.~Ding, W.~Liu, H.~Piao, Y.~Wang, X.~Yang, and B.~Yin, ``A
  vision-based irregular obstacle avoidance framework via deep reinforcement
  learning,'' in \emph{Proc. IEEE/RSJ Int. Conf. Intell. Robots Syst.}\hskip
  1em plus 0.5em minus 0.4em\relax IEEE, 2021, pp. 9262--9269.

\bibitem{geiger2013vision}
A.~Geiger, P.~Lenz, C.~Stiller, and R.~Urtasun, ``Vision meets robotics: The
  kitti dataset,'' \emph{Int. J. Robot. Res.}, vol.~32, no.~11, pp. 1231--1237,
  2013.

\bibitem{giusti2015machine}
A.~Giusti, J.~Guzzi, D.~C. Cire{\c{s}}an, F.-L. He, J.~P. Rodr{\'\i}guez,
  F.~Fontana, M.~Faessler, C.~Forster, J.~Schmidhuber, G.~Di~Caro,
  \emph{et~al.}, ``A machine learning approach to visual perception of forest
  trails for mobile robots,'' \emph{IEEE Robot. Autom. Lett.}, vol.~1, no.~2,
  pp. 661--667, 2015.

\bibitem{gordon2019splitnet}
D.~Gordon, A.~Kadian, D.~Parikh, J.~Hoffman, and D.~Batra, ``Splitnet: Sim2sim
  and task2task transfer for embodied visual navigation,'' in \emph{Proc. IEEE
  Int. Conf. Comput. Vis.}, 2019, pp. 1022--1031.

\bibitem{hirose2019deep}
N.~Hirose, F.~Xia, R.~Mart{\'\i}n-Mart{\'\i}n, A.~Sadeghian, and S.~Savarese,
  ``Deep visual mpc-policy learning for navigation,'' \emph{IEEE Robot. Autom.
  Lett.}, vol.~4, no.~4, pp. 3184--3191, 2019.

\bibitem{hochreiter1997long}
S.~Hochreiter and J.~Schmidhuber, ``Long short-term memory,'' \emph{Neural
  Comput.}, vol.~9, no.~8, pp. 1735--1780, 1997.

\bibitem{howard2017mobilenets}
A.~G. Howard, M.~Zhu, B.~Chen, D.~Kalenichenko, W.~Wang, T.~Weyand,
  M.~Andreetto, and H.~Adam, ``Mobilenets: Efficient convolutional neural
  networks for mobile vision applications,'' \emph{arXiv:1704.04861}, 2017.

\bibitem{huang2004global}
Y.-W. Huang, S.-Y. Chien, B.-Y. Hsieh, and L.-G. Chen, ``Global elimination
  algorithm and architecture design for fast block matching motion
  estimation,'' \emph{IEEE Transactions on circuits and systems for Video
  technology}, vol.~14, no.~6, pp. 898--907, 2004.

\bibitem{huang2019ccnet}
Z.~Huang, X.~Wang, L.~Huang, C.~Huang, Y.~Wei, and W.~Liu, ``Ccnet: Criss-cross
  attention for semantic segmentation,'' in \emph{Proc. IEEE Int. Conf. Comput.
  Vis.}, 2019, pp. 603--612.

\bibitem{hutchinson1996tutorial}
S.~Hutchinson, G.~D. Hager, and P.~I. Corke, ``A tutorial on visual servo
  control,'' \emph{IEEE Trans. Robot. Autom.}, vol.~12, no.~5, pp. 651--670,
  1996.

\bibitem{kim2015deep}
D.~K. Kim and T.~Chen, ``Deep neural network for real-time autonomous indoor
  navigation,'' \emph{arXiv:1511.04668}, 2015.

\bibitem{lengwehasarit2001probabilistic}
K.~Lengwehasarit and A.~Ortega, ``Probabilistic partial-distance fast matching
  algorithms for motion estimation,'' \emph{IEEE Transactions on Circuits and
  Systems for Video Technology}, vol.~11, no.~2, pp. 139--152, 2001.

\bibitem{li2020hrl4in}
C.~Li, F.~Xia, R.~Martin-Martin, and S.~Savarese, ``Hrl4in: Hierarchical
  reinforcement learning for interactive navigation with mobile manipulators,''
  in \emph{Conference on Robot Learning}.\hskip 1em plus 0.5em minus
  0.4em\relax PMLR, 2020, pp. 603--616.

\bibitem{Tianyu2018}
T.~Liu, ``Robot collision avoidance via deep reinforcement learning,''
  \url{https://github.com/Acmece/rl-collision-avoidance.git}, 2018.

\bibitem{long2018towards}
P.~Long, T.~Fanl, X.~Liao, W.~Liu, H.~Zhang, and J.~Pan, ``Towards optimally
  decentralized multi-robot collision avoidance via deep reinforcement
  learning,'' in \emph{Proc. IEEE Int. Conf. Robot. Autom.}, 2018, pp.
  6252--6259.

\bibitem{lu2021mgrl}
Y.~Lu, Y.~Chen, D.~Zhao, and D.~Li, ``Mgrl: graph neural network based
  inference in a markov network with reinforcement learning for visual
  navigation,'' \emph{Neurocomputing}, vol. 421, pp. 140--150, 2021.

\bibitem{ma2019learning}
L.~Ma, Y.~Liu, J.~Chen, and D.~Jin, ``Learning to navigate in indoor
  environments: from memorizing to reasoning,'' \emph{arXiv:1904.06933}, 2019.

\bibitem{mccarthy2004performance}
C.~McCarthy and N.~Bames, ``Performance of optical flow techniques for indoor
  navigation with a mobile robot,'' in \emph{Proc. IEEE Int. Conf. Robot.
  Autom.}, vol.~5, 2004, pp. 5093--5098.

\bibitem{michels2005high}
J.~Michels, A.~Saxena, and A.~Y. Ng, ``High speed obstacle avoidance using
  monocular vision and reinforcement learning,'' in \emph{Proc. Int. Conf.
  Mach. Learn.}, 2005, pp. 593--600.

\bibitem{mur2015orb}
R.~Mur-Artal, J.~M.~M. Montiel, and J.~D. Tardos, ``Orb-slam: a versatile and
  accurate monocular slam system,'' \emph{IEEE Trans. Robot.}, vol.~31, no.~5,
  pp. 1147--1163, 2015.

\bibitem{nguyen2007real}
T.~H. Nguyen, J.~S. Nguyen, D.~M. Pham, and H.~T. Nguyen, ``Real-time obstacle
  detection for an autonomous wheelchair using stereoscopic cameras,'' in
  \emph{2007 29th Annual International Conference of the IEEE Engineering in
  Medicine and Biology Society}.\hskip 1em plus 0.5em minus 0.4em\relax IEEE,
  2007, pp. 4775--4778.

\bibitem{okal2014towards}
B.~Okal and K.~O. Arras, ``Towards group-level social activity recognition for
  mobile robots,'' in \emph{IROS Assistance and Service Robotics in a Human
  Environments Workshop}, 2014.

\bibitem{peng2018sim}
X.~B. Peng, M.~Andrychowicz, W.~Zaremba, and P.~Abbeel, ``Sim-to-real transfer
  of robotic control with dynamics randomization,'' in \emph{Proc. IEEE Int.
  Conf. Robot. Autom.}, 2018, pp. 1--8.

\bibitem{pfeiffer2017perception}
M.~Pfeiffer, M.~Schaeuble, J.~Nieto, R.~Siegwart, and C.~Cadena, ``From
  perception to decision: A data-driven approach to end-to-end motion planning
  for autonomous ground robots,'' in \emph{Proc. IEEE Int. Conf. Robot.
  Autom.}, 2017, pp. 1527--1533.

\bibitem{savva2019habitat}
M.~Savva, A.~Kadian, O.~Maksymets, Y.~Zhao, E.~Wijmans, B.~Jain, J.~Straub,
  J.~Liu, V.~Koltun, J.~Malik, \emph{et~al.}, ``Habitat: A platform for
  embodied ai research,'' in \emph{Proc. IEEE Int. Conf. Comput. Vis.}, 2019,
  pp. 9339--9347.

\bibitem{schroeder2012towards}
S.~Schroeder, M.~Zilske, G.~Liedtke, and K.~Nagel, ``Towards a multi-agent
  logistics and commercial transport model: The transport service provider's
  view,'' \emph{Procedia-Social and Behavioral Sciences}, vol.~39, pp.
  649--663, 2012.

\bibitem{schulman2017proximal}
J.~Schulman, F.~Wolski, P.~Dhariwal, A.~Radford, and O.~Klimov, ``Proximal
  policy optimization algorithms,'' \emph{arXiv:1707.06347}, 2017.

\bibitem{silberman2012indoor}
N.~Silberman, D.~Hoiem, P.~Kohli, and R.~Fergus, ``Indoor segmentation and
  support inference from rgbd images,'' in \emph{Proc. Eur. Conf. Comput.
  Vision.}, 2012, pp. 746--760.

\bibitem{souhila2007optical}
K.~Souhila and A.~Karim, ``Optical flow based robot obstacle avoidance,''
  \emph{Int. J. Adv. Robot. Syst.}, vol.~4, no.~1, p.~2, 2007.

\bibitem{tai2016deep}
L.~Tai, S.~Li, and M.~Liu, ``A deep-network solution towards model-less
  obstacle avoidance,'' in \emph{Proc. IEEE/RSJ Int. Conf. Intell. Robots
  Syst.}, 2016, pp. 2759--2764.

\bibitem{tai2017virtual}
L.~Tai, G.~Paolo, and M.~Liu, ``Virtual-to-real deep reinforcement learning:
  Continuous control of mobile robots for mapless navigation,'' in \emph{Proc.
  IEEE/RSJ Int. Conf. Intell. Robots Syst.}, 2017, pp. 31--36.

\bibitem{tan2018sim}
J.~Tan, T.~Zhang, E.~Coumans, A.~Iscen, Y.~Bai, D.~Hafner, S.~Bohez, and
  V.~Vanhoucke, ``Sim-to-real: Learning agile locomotion for quadruped
  robots,'' \emph{arXiv:1804.10332}, 2018.

\bibitem{tang2020reinforcement}
G.~Tang, N.~Kumar, and K.~P. Michmizos, ``Reinforcement co-learning of deep and
  spiking neural networks for energy-efficient mapless navigation with
  neuromorphic hardware,'' in \emph{Proc. IEEE/RSJ Int. Conf. Intell. Robots
  Syst.}\hskip 1em plus 0.5em minus 0.4em\relax IEEE, 2020, pp. 6090--6097.

\bibitem{tobin2017domain}
J.~Tobin, R.~Fong, A.~Ray, J.~Schneider, W.~Zaremba, and P.~Abbeel, ``Domain
  randomization for transferring deep neural networks from simulation to the
  real world,'' in \emph{Proc. IEEE/RSJ Int. Conf. Intell. Robots Syst.}, 2017,
  pp. 23--30.

\bibitem{van2011reciprocal}
J.~Van Den~Berg, S.~J. Guy, M.~Lin, and D.~Manocha, ``Reciprocal n-body
  collision avoidance,'' in \emph{Robot. Res.}, 2011, pp. 3--19.

\bibitem{vanne2006high}
J.~Vanne, E.~Aho, T.~D. Hamalainen, and K.~Kuusilinna, ``A high-performance sum
  of absolute difference implementation for motion estimation,'' \emph{IEEE
  transactions on circuits and systems for video technology}, vol.~16, no.~7,
  pp. 876--883, 2006.

\bibitem{wang2021modular}
J.~Wang, S.~Elfwing, and E.~Uchibe, ``Modular deep reinforcement learning from
  reward and punishment for robot navigation,'' \emph{Neural Networks}, vol.
  135, pp. 115--126, 2021.

\bibitem{wang2019pseudo}
Y.~Wang, W.-L. Chao, D.~Garg, B.~Hariharan, M.~Campbell, and K.~Q. Weinberger,
  ``Pseudo-lidar from visual depth estimation: Bridging the gap in 3d object
  detection for autonomous driving,'' in \emph{Proc. IEEE Conf. Comput. Vis.
  Pattern Recognit.}, 2019.

\bibitem{wofk2019fastdepth}
D.~Wofk, F.~Ma, T.-J. Yang, S.~Karaman, and V.~Sze, ``Fastdepth: Fast monocular
  depth estimation on embedded systems,'' in \emph{Proc. IEEE Int. Conf. Robot.
  Autom.}, 2019, pp. 6101--6108.

\bibitem{wu2020neonav}
Q.~Wu, D.~Manocha, J.~Wang, and K.~Xu, ``Neonav: Improving the generalization
  of visual navigation via generating next expected observations,'' in
  \emph{Proceedings of the AAAI Conference on Artificial Intelligence},
  vol.~34, no.~06, 2020, pp. 10\,001--10\,008.

\bibitem{wu2018building}
Y.~Wu, Y.~Wu, G.~Gkioxari, and Y.~Tian, ``Building generalizable agents with a
  realistic and rich 3d environment,'' \emph{arXiv:1801.02209}, 2018.

\bibitem{wu2019exploring}
Y.~Wu, Z.~Rao, W.~Zhang, S.~Lu, W.~Lu, and Z.-J. Zha, ``Exploring the task
  cooperation in multi-goal visual navigation.'' in \emph{IJCAI}, 2019, pp.
  609--615.

\bibitem{yang2018visual}
W.~Yang, X.~Wang, A.~Farhadi, A.~Gupta, and R.~Mottaghi, ``Visual semantic
  navigation using scene priors,'' \emph{arXiv preprint arXiv:1810.06543},
  2018.

\bibitem{yang2019my}
X.~Yang, H.~Mei, K.~Xu, X.~Wei, B.~Yin, and R.~W. Lau, ``Where is my mirror?''
  in \emph{Proc. IEEE Int. Conf. Comput. Vis.}, 2019, pp. 8809--8818.

\bibitem{zhang2020language}
W.~Zhang, C.~Ma, Q.~Wu, and X.~Yang, ``Language-guided navigation via
  cross-modal grounding and alternate adversarial learning,'' \emph{IEEE
  Transactions on Circuits and Systems for Video Technology}, 2020.

\bibitem{zhu2017target}
Y.~Zhu, R.~Mottaghi, E.~Kolve, J.~J. Lim, A.~Gupta, L.~Fei-Fei, and A.~Farhadi,
  ``Target-driven visual navigation in indoor scenes using deep reinforcement
  learning,'' in \emph{Proc. IEEE Int. Conf. Robot. Autom.}, 2017, pp.
  3357--3364.

\end{thebibliography}

\clearpage

\begin{IEEEbiography}[{\includegraphics[width=1in,height=1.25in,clip,keepaspectratio]{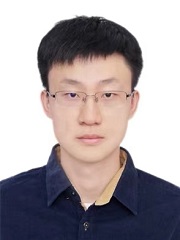}}]{Jianchuan Ding}
	received his M.Sc. degree in the School of Computer Science and Technology, Dalian University of Technology, Dalian, China, in 2022, and his B.Sc. degree from Liaoning University, Shenyang, China, in 2019. He is currently working in Dalian University of Technology and Hebei University of Water Resources and Electric Engineering. His research interests include robotic science and computer vision.
\end{IEEEbiography}

\begin{IEEEbiography}[{\includegraphics[width=1.25in,height=1.5in,clip,keepaspectratio]{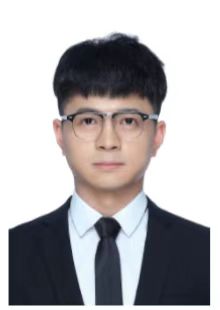}}]{Lingping Gao}
	is an engineer in Alibaba Damo Academy. He received his M.Sc. degree in the School of Computer Science and Technology, Dalian University of Technology, Dalian, China, in 2020, and his B.Sc. degree from Qingdao University of Science and Technology, Qingdao, China, in 2018. His main research interests include deep reinforcement learning, path planning, computer vision and so on.
\end{IEEEbiography}

\begin{IEEEbiography}[{\includegraphics[width=1in,height=1.25in,clip,keepaspectratio]{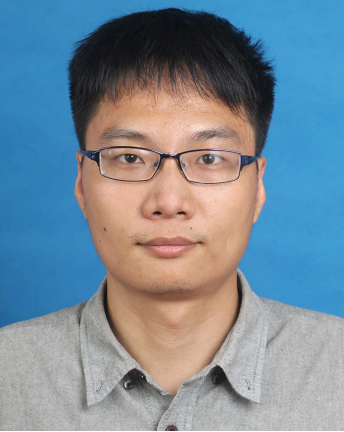}}]{Wenxi Liu}
	received the Ph.D. degree from the City University of Hong Kong. He is currently an Associate Professor with the College of Mathematics and Computer Science, Fuzhou University. His research interests include computer vision, robot vision, and image processing.
\end{IEEEbiography}

\begin{IEEEbiography}[{\includegraphics[width=1in,height=1.25in,clip]{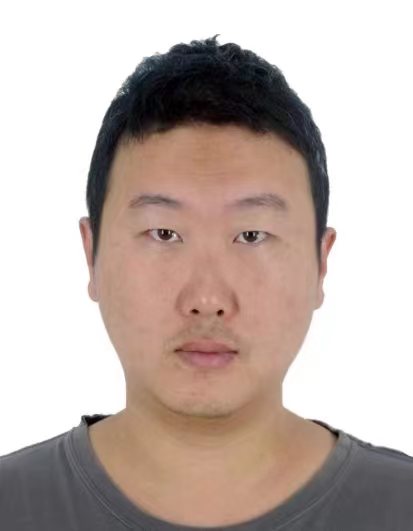}}]{Haiyin Piao} is currently working toward the Ph.D. degree in School of Electronics and Information, Northwestern Polytechnical University, China. He is also the vice manager of AI center, SADRI institute, China. He received the M.Sc. degree in computer science from Dalian University of Technology, China, in 2010. Recently, he has published more than 30 articles in international journals and conferences, including NeurlPS, ICAPS, IEEE TITS, IEEE TCSVT, IEEE TETCI, EAAI, etc. His current research interests include deep learning, multiagent reinforcement learning, and game theory with particular attention to aerospace applications.
\end{IEEEbiography}

\begin{IEEEbiography}[{\includegraphics[width=1in,height=1.25in,clip,keepaspectratio]{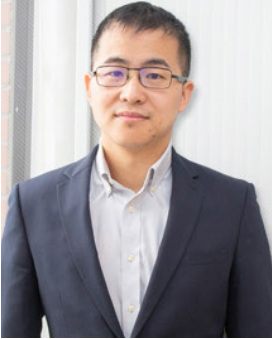}}]{Jia Pan}
	received the B.S. degree in control theory and engineering from Tsinghua University, in 2005, the M.S. degree from the Chinese Academy of Sciences, in 2008, where he worked on Computer-Aided Design (CAD), and the Ph.D. degree from the Department of Computer Science, University of North Carolina at Chapel Hill (UNC), in 2013. In 2014, he completed a postdoctoral position at UC Berkeley, working with Pieter Abbeel. In October 2014, he started at the Computer Science Department, University of Hong Kong, and then moved to the Department of Mechanical and Biomedical Engineering, City University of Hong Kong, as an Assistant Professor, in 2015. After three years, he moved back to The University of Hong Kong, where he is currently an Assistant Professor with the Computer Science Department. His research focuses on creating algorithms that allow robots to efficiently and intelligently interact with the world and collaborate with people. These general-purpose sensing, control, planning, and manipulation algorithms can be applied to robots that work in homes, factories, laboratories, or fields.
\end{IEEEbiography}

\begin{IEEEbiography}[{\includegraphics[width=1in,height=1.25in,clip,keepaspectratio]{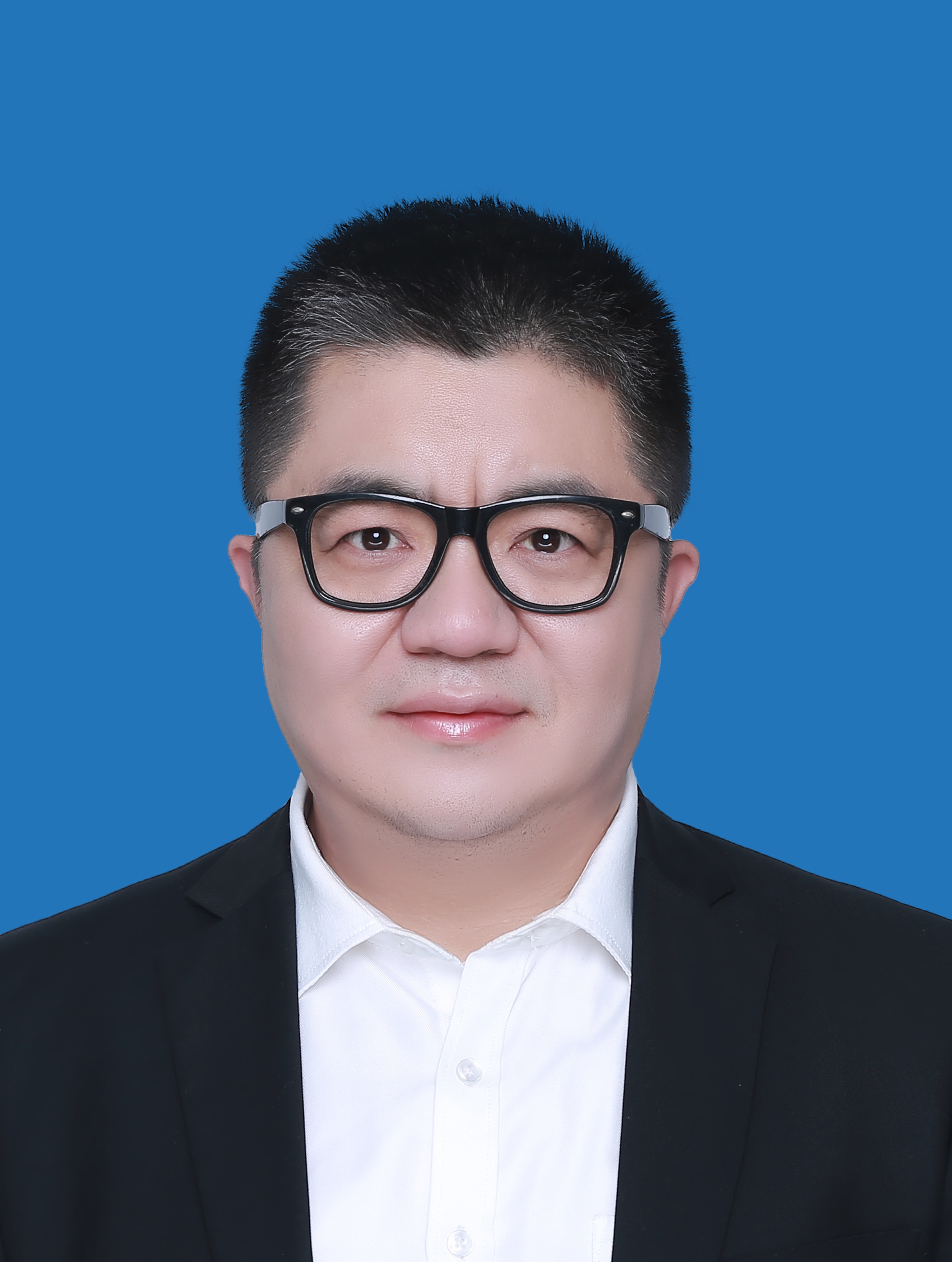}}]{Zhenjun Du}
has a doctor's degree in pattern recognition and intelligent system of Shenyang Institute of automation, Chinese Academy of Sciences. He is currently the Technical Executive Director and President of Central Research Institute of Shenyang SIASUN Robot Automation Co., Ltd.
\end{IEEEbiography}

\begin{IEEEbiography}[{\includegraphics[width=1in,height=1.25in,clip,keepaspectratio]{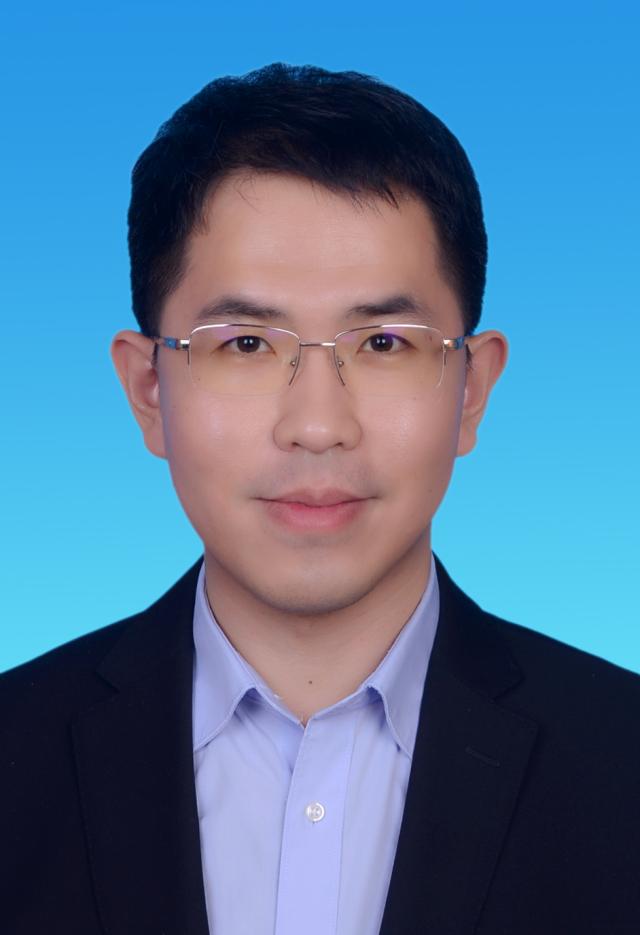}}]{Xin Yang}
	is a professor and doctoral advisor at Dalian University of Technology. He received his Ph.D. degree in computer science from Zhejiang University (2007-2012), and his B.S. degree in computer science from Jilin University (2003-2007). His main research interests include computer graphics and vision, intelligent robot technology, focusing on the efficient expression, understanding, perception and interaction of scenes.
\end{IEEEbiography}

\begin{IEEEbiography}[{\includegraphics[width=1in,height=1.25in,clip,keepaspectratio]{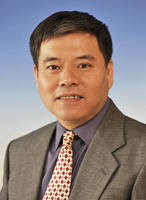}}]{Baocai Yin}
	is a professor and doctoral advisor at Dalian University of Technology. He received his Ph.D. degree in computational mathematics from Dalian University of Technology (1990-1993), where he also received his M.S. degree in computational mathematics (1985-1988) and his B.S. degree in applied mathematics(1981-1985). His research areas include digital multimedia technology, virtual reality and graphics technology, and multi-function perception technology.
\end{IEEEbiography}

%








\end{document}